\documentclass[10pt,twocolumn,letterpaper]{article}

\usepackage{cvpr}              

\usepackage{graphicx}
\usepackage{amsmath}
\usepackage{amssymb}
\usepackage{booktabs}
\usepackage{xcolor}
\usepackage{tikz}
\usetikzlibrary{mindmap} 
\usepackage{arydshln}

\usepackage[accsupp]{axessibility} 

\definecolor{JINZAMOMI}{RGB}{225,122,119} 
\definecolor{AKEBONO}{RGB}{241,148,131} 
\definecolor{TOKI}{RGB}{238,169,169} 
\definecolor{MOGEI}{RGB}{123,162,63} 
\definecolor{HIWA}{RGB}{190,194,63} 
\definecolor{NOSHIMEHANA}{RGB}{43,95,117}
\definecolor{KUCHINASHI}{RGB}{246,197,85} 
\definecolor{USUKI}{RGB}{250,214,137} 
\definecolor{HANAASAGI}{RGB}{30,136,168} 
\definecolor{SORA}{RGB}{88,178,220} 
\definecolor{NAE}{RGB}{134,193,102} 
\definecolor{KOKE}{RGB}{131,138,45} 
\definecolor{WASURENAGUSA}{RGB}{125,185,222} 
\definecolor{GUNJYO}{RGB}{81,168,221} 
\definecolor{KAMENOZOKI}{RGB}{165,222,228} 

\usepackage[pagebackref,breaklinks,colorlinks]{hyperref}

\usepackage[capitalize]{cleveref}
\crefname{section}{Sec.}{Secs.}
\Crefname{section}{Section}{Sections}
\Crefname{table}{Table}{Tables}
\crefname{table}{Tab.}{Tabs.}

\usepackage{multirow}
\usepackage{fontawesome}
\usepackage{colortbl}

\usepackage{tcolorbox}
\usepackage{xpatch}
\usepackage{diagbox}
\usepackage{utfsym}

\usepackage{amsmath}
\usepackage{amsfonts}
\usepackage{amssymb}
\usepackage{mathrsfs}

\usepackage[T1]{fontenc} 
\pdfmapline{+delphine < Delphine.ttf <T1-WGL4.enc}

\usepackage{MnSymbol}
\usepackage{makecell}

\def\cvprPaperID{49} 
\def\confName{CVPR}
\def\confYear{2023}

\begin{document}

\title{Foundation Models for Biomedical Image Segmentation: A Survey}


\author{
Ho Hin Lee*, Yu Gu*, Theodore Zhao, Yanbo Xu, Jianwei Yang, Naoto Usuyama, Cliff Wong, Mu Wei, \\ 
Bennett A. Landman, Yuankai Huo, Alberto Santamaria-Pang, Hoifung Poon\\
Microsoft Research, Vanderbilt University\\
}

\maketitle

\begin{abstract}

Recent advancements in biomedical image analysis have been significantly driven by the Segment Anything Model (SAM). This transformative technology, originally developed for general-purpose computer vision, has found rapid application in medical image processing. Within the last year, marked by over 100 publications, SAM has demonstrated its prowess in zero-shot learning adaptations for medical imaging. The fundamental premise of SAM lies in its capability to segment or identify objects in images without prior knowledge of the object type or imaging modality. This approach aligns well with tasks achievable by the human visual system, though its application in non-biological vision contexts remains more theoretically challenging. A notable feature of SAM is its ability to adjust segmentation according to a specified resolution scale or area of interest, akin to semantic priming. This adaptability has spurred a wave of creativity and innovation in applying SAM to medical imaging. Our review focuses on the period from April 1, 2023, to September 30, 2023, a critical first six months post-initial publication. We examine the adaptations and integrations of SAM necessary to address longstanding clinical challenges, particularly in the context of 33 open datasets covered in our analysis. While SAM approaches or achieves state-of-the-art performance in numerous applications, it falls short in certain areas, such as segmentation of the carotid artery, adrenal glands, optic nerve, and mandible bone. Our survey delves into the innovative techniques where SAM's foundational approach excels and explores the core concepts in translating and applying these models effectively in diverse medical imaging scenarios.

\end{abstract}


\section{Introduction}
In the rapidly evolving landscape of biomedical image analysis, the Segment Anything Model (SAM) has emerged as a paradigm-shifting technology. Initially conceived for general-purpose computer vision, SAM has quickly become instrumental in the field of medical image processing. This survey provides an in-depth examination of SAM's applications and adaptations in biomedical imaging, focusing on its remarkable proliferation in scholarly research over the past year. \\
One of the most compelling attributes of SAM is its ability to segment objects in medical images without requiring prior knowledge about the object types or imaging modalities. This capability is of paramount importance in biomedical imaging, where precise segmentation of diverse anatomical structures and pathologies is critical. SAM's methodology echoes the adaptability of the human visual system in object recognition and segmentation, marking a significant advancement in computational image analysis. \\
Our review concentrates on the period from April 1, 2023, to September 30, 2023. These six months post-SAM's initial publication represent a crucial phase in its development, characterized by extensive adaptations and integrations of SAM to meet longstanding clinical challenges, especially in the context of open datasets. Our analysis encompasses 33 such datasets, offering a comprehensive assessment of the cutting-edge applications of SAM in this dynamic area. \\
Moreover, we provide a critical evaluation of the domains where SAM's performance is currently lacking, including the segmentation of complex anatomical regions such as the carotid artery, adrenal glands, optic nerve, and mandible bone. This investigation aims to highlight the current limitations and identify potential areas for future enhancements in SAM technology. \\
This survey also explores the innovative techniques where SAM’s fundamental approach demonstrates exceptional performance. We delve into the underlying concepts essential for the effective translation and application of these models across various medical imaging scenarios. Our objective is to delineate the current state of SAM applications in biomedical imaging and to offer strategic directions for future research in this vibrant and rapidly advancing field.\\
In summary, as SAM propels forward in the realm of biomedical imaging, its multifaceted impact becomes increasingly evident. Beyond the technical prowess in image segmentation, SAM represents a beacon of interdisciplinary collaboration, merging the frontiers of computer science, medicine, and data analytics. This survey highlights not just SAM's technical achievements but also its role as a catalyst in redefining medical research paradigms. By examining the integration of SAM across various medical domains, we offer insights into how this technology is reshaping diagnostic methodologies, enhancing patient care, and opening new avenues in personalized medicine. Our exploration into SAM's application in diverse medical contexts is not just an academic exercise but a testament to the transformative potential of AI in healthcare, providing a roadmap for future innovations that could further revolutionize the field.

\section{Background} 
\subsection{Segment Anything Model (SAM)}

Building upon foundational model concepts, the Segment Anything Model (SAM) \cite{kirillov2023segment} serves as a pioneering framework that facilitates the expansion of segmentation capabilities to previously unseen objects, leveraging user-provided prompt guidance. Prompt inputs, in their diverse forms--ranging from singular or multiple points to bounding boxes, or text descriptions \cite{zou2023segment}--empower SAM to craft accurate instance segmentation masks for new image samples without necessitating further training (a process known as zero-shot segmentation). It should be highlighted, however, that the segmented objects aren't imbued with semantic meaning, a characteristic contingent upon the reliability of the prompts utilized.

Structurally, SAM's architecture can be divided into three primary components: \\
\indent\textbf{Image Encoder.} This component incorporates a Masked AutoEncoder (MAE) \cite{he2022masked} pre-trained Vision Transformer (ViT) \cite{dosovitskiy2020image} to adapt to high-resolution images across various scales. In order to explore scalability, the image encoder is modulated from ViT-B to ViT-H or other model architectures, allowing for an assessment of potential trade-offs in model efficiency. All images are resized to dimensions of $1024\times 1024$ to conform to the standardized image size established during pre-training.\\
\indent\textbf{Prompt Encoder.} Two primary categories of prompts are delineated for implementation: sparse and dense prompts. Sparse prompts encompass elements like points and boxes, which are processed using ViT's positional encoding concept, integrated with their respective learned representations. Additionally, text forms a part of the sparse prompt category, with features extracted through a CLIP-pretrained text encoder. Meanwhile, dense prompts, typified by masks, undergo convolution operations to establish element-wise correlations with image features discerned by the image encoder.\\

\indent\textbf{Light-weighted Mask Decoder} This segment aims to unify features discerned by various encoders, fostering detailed predictions. It incorporates a transformer decoder block followed by a dynamic prediction head, adept at merging bidirectional self-attention and cross-attention mechanisms, thereby facilitating concurrent updates of prompt-to-image and image-to-prompt embeddings. To maintain computational efficiency, only two decoder blocks are deployed to upsample the resolution of the acquired embeddings. A Multi-Layer Perceptron (MLP) is utilized as the final layer that maps output tokens to a dynamic linear classifier and ascertains foreground probabilities essential for segmentation.\\
The entire SAM is trained on the large-scale dataset SA-1B, which consists of 11M high-resolution images with 1.1B high-quality segmentation masks, 400× more masks than any existing segmentation dataset \cite{kirillov2023segment}. Such large-scale pre-training enhances both the feasibility and stability of adapting unseen domain samples for segmentation. While its success in the general domain is noteworthy, medical imaging presents its own set of intricacies.
\begin{figure*}[tb]

\centering
\includegraphics[width=0.8\linewidth]{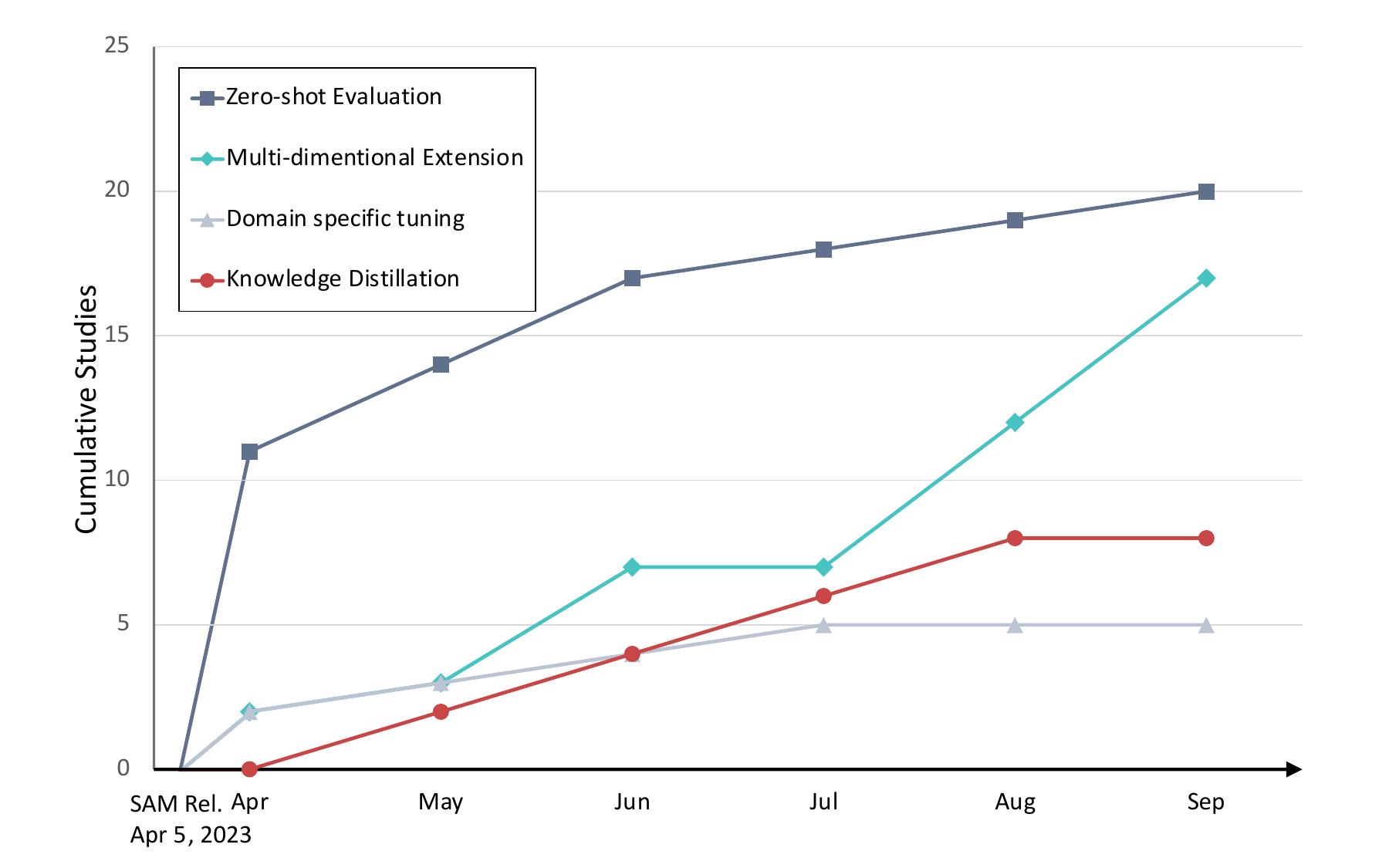}
\caption{\textbf{Evolution of SAM's adaptation in medical research from April to September 2023.} The graph showcases the cumulative studies emphasizing four phases: (i) Zero-shot Evaluation, (ii) Multi-dimensional Extension, (iii) Domain-specific Tuning, and (iv) Knowledge Distillation, highlighting a growing research interest in optimizing SAM for medical image segmentation.}
\label{fig:timeline}
\end{figure*}
\subsection{Medical Image Segmentation}

Medical image segmentation stands as an indispensable tool in the clinical realm, enabling the precise extraction of anatomical structures and pathological features from a plethora of imaging modalities. It's instrumental in facilitating accurate diagnoses, formulating treatment plans, guiding surgical interventions, and propelling medical research. 

\subsubsection{Data}
As medical images reconstructed with multi-dimensional property (i.e., 2D, 3D) due to the significant variability in imaging protocols, manually generating segmentation mask for organ-specific regions are time-consuming with significant efforts from clinicians. Initial efforts have been put to generate datasets with single organ labeled ~\cite{clark2013cancer_TCIA, roth2014new, Roth2015, roth2015deeporgan, roth2016data}. The single organ labeled datasets provide an initial foundation basis to train deep learning models for automatic segmentation, thus enhancing the extensive capabilities to label multiple organ regions. The capability of deep learning models for directly segmenting multiple regions also starts to explore with multi-organ/multi-tissue labeled datasets \cite{landman2015miccai, bilic2023liver, ACDCData, antonelli2022medical}. However, as shown in Table \ref{tab:3d-dataset}, only limited number of semantic classes (i.e. 15 organs) have demonstrated across different datasets \cite{ji2022amos}, unlike SA-1B with large semantic classes knowledge. TotalSegmentator dataset is a significant breakthrough with 117 organs labeled by adapting hierarchical segmentation network nnUNet and refining the pseudo label manually. Such dataset demonstrates similarity with SA-1B with approximately 100 semantic classes for each 2D slice-wise image. However, with the significant manual efforts contributed in labeling, only 1228 imaging samples are labeled and there is still a significant difference in sampling sizes comparing to natural image dataset.


\subsubsection{Methods}
\textbf{Traditional Techniques.} Grounded in algorithmic methods, early approaches like thresholding \cite{feng2017multi}, region-growing \cite{mubarak2012hybrid}, and contour tracing provided rudimentary segmentation \cite{xu2000image, shen2010active}. These were often manual or semi-automatic, rendering them labor-intensive and sensitive to variations. 

\textbf{Convolutional Neural Networks.} The rise of machine learning witnessed the advent of CNNs, with architectures like U-Net revolutionizing medical image segmentation\cite{ronneberger2015u, cciccek20163d, isensee2021nnu, lee20223d, lee2023scaling}. By hierarchically processing images, CNNs captured both local features and broader structures, albeit sometimes grappling with global context due to fixed receptive fields.

\textbf{Vision Transformers.} Emerging as a recent trend, ViTs offer the capability to understand global image patterns. Unlike CNNs, they aren't bound by fixed receptive fields, enabling them to capture both local details and broader contexts. Hierarchical transformers, which blend CNN and transformer strengths, further refine segmentation capabilities\cite{chen2021transunet, xie2021cotr, wang2021transbts, zhou2021nnformer, hatamizadeh2022unetr, hatamizadeh2022swin}.



\subsection{SAM's Role in Medical Image Segmentation}
The application of SAM, initially optimized for general images, to the specialized realm of medical imaging introduces both opportunities and barriers. The following are pivotal considerations:

\textbf{Domain Specificity.} SAM's general-domain proficiency is laudable, but medical images introduce nuances not often seen in everyday images. Variations due to patient demographics, imaging equipment, and protocols add layers of complexity \cite{mazurowski2023segment, wu2023medical, ma2023segment}. Such diversities question whether SAM, in its original state, can deliver consistent performance across the vast spectrum of medical images. Domain-specific fine-tuning and pretraining, incorporating expert annotations or domain-adaptive techniques, may be needed for optimal performance\cite{Gu_2021}.

\textbf{Dimenstional Mismatch.} While 2D images are commonplace, medical imaging often delves into the third dimension. Modalities like MRI and CT offer volumetric insights, crucial for understanding anatomical structures and pathologies\cite{lee20223d, hatamizadeh2022swin}, while SAM's 2D-centric design is not natively cater to such data. Adapting it would likely necessitate architectural tweaks and potential integration with 3D convolutional layers or volumetric transformers.

\textbf{Data Scarcity and Annotation Quality.} Accurate segmentation hinges on high-quality annotations, which, in the medical domain, are time-consuming, expertise-dependent, and fraught with privacy concerns\cite{zhang2023samdsk}. SAM's broad training on general images offers a potential springboard, but its true utility in addressing the challenges of limited medical data and annotation variance remains an open question.

\section{Challenges in Contemporary Medical Image Segmentation Tasks}

To understand the status of contemporary medical image segmentation, we curated a collection of 33 medical image segmentation datasets in Table \ref{tab:t2i-dataset}. We order the datasets by the their most recent release year, ranging from 1998 to 2023, with the majority of them in the past decade. All the datasets in Table \ref{tab:t2i-dataset} are publicly available, thus have direct impact on the development of open source models. We cover a variety of 7 modalities, 6 segmentation target types, 17 anatomies in this collection. Although it is impossible to exhaust every medical datasets that are publicly available, we view this collection as a representative sub-sample to reflect the broad medical image segmentation problems from a high level.

In the following parts of this section, we analyze the challenges of medical image segmentation from two aspects: 1. input (imaging modality) and 2. output (segmentation target), showing their versatility and how they are different from the problems in general domain.

\begin{table*}[h!]
\centering
\fontsize{7}{9.3}\selectfont
\caption{\textbf{Chronological timeline of medical image segmentation datasets.} \textit{“Public”} includes a link to each dataset (if available) or paper (if not). \textit{“Annotations”} denotes the number of classes with ground-truth quality labels in each dataset.}
\begin{tabular}{cccccccccc}
\toprule[1.5pt]
 \multirow{2.5}{*}{\small \textbf{Year}} &\multirow{2.5}{*}{\small \textbf{Dataset}} & \multirow{2.5}{*}{\small \textbf{Public}} & \multicolumn{6}{c}{\small \textbf{Details}} \\ \cmidrule{4-9}
                              &              &       &  Modality    & Anatomy  & Data Size & Label Quality & \# Targets & Seg. Target Type              \\ \hline

\cellcolor{HANAASAGI!3} 1998 & JSRT~\cite{shiraishi2000development_JSRT} & \href{http://imgcom.jsrt.or.jp/minijsrtdb/}{\color{green}\faCheckCircle} & X-Ray & Chest
 & 307 & Manual & 2 & Multi-Organ \\

\cellcolor{HANAASAGI!6} 2012 & VESSEL12~\cite{rudyanto2014comparing_VESSEL12} & \href{https://vessel12.grand-challenge.org/}{\color{green}\faCheckCircle} & CT & Lung & 20 & Manual & 1 & Organ Parts \\
                              
\cellcolor{HANAASAGI!9} 2012&  PROMISE12~\cite{Promise12Data} & \href{https://promise12.grand-challenge.org/}{\color{green}\faCheckCircle}  & MRI & Prostate & 100 & Manual & 1 & Single Organ \\

\cellcolor{HANAASAGI!12} 2013 & NCI-ISBI~\cite{Bloch2015_NCI-ISBI2013} & \href{https://wiki.cancerimagingarchive.net/pages/viewpage.action?pageId=21267207}{\color{green}\faCheckCircle} & MRI & Prostate
 & 80 & Manual & 2 & Organ Parts \\


\cellcolor{HANAASAGI!18} 2015&  BTCV~\cite{landman2015miccai} & \href{https://www.synapse.org/#!Synapse:syn3193805}{\color{green}\faCheckCircle}  & CT & Abdomen & 50 & Manual & 13 & Multi-Organ  \\

\cellcolor{HANAASAGI!21} 2015 & CT-Lymph Nodes~\cite{Roth2015, roth2014new,clark2013cancer_TCIA} &  \href{https://wiki.cancerimagingarchive.net/pages/viewpage.action?pageId=19726546}{\color{green}\faCheckCircle} & CT & Mediastinum & 176 & Manual & 1 & Single Organ \\

\cellcolor{HANAASAGI!21} 2015 & GlaS~\cite{sirinukunwattana2016gland, glas} &  \href{https://warwick.ac.uk/fac/cross_fac/tia/data/glascontest/}{\color{green}\faCheckCircle} & Pathology & Colon & 165 & Manual & 1 & Cells \\

\cellcolor{HANAASAGI!21} 2016 & Pancreas-CT~\cite{roth2016data,roth2015deeporgan,clark2013cancer_TCIA} &  \href{https://wiki.cancerimagingarchive.net/display/public/pancreas-ct#22514040c56be89073824ef7946e58d813146283/}{\color{green}\faCheckCircle} & CT & Pancreas & 80 & Manual & 1 & Single Organ \\

\cellcolor{HANAASAGI!24} 2017& LiTS~\cite{bilic2023liver} & \href{https://competitions.codalab.org/competitions/17094}{\color{green}\faCheckCircle} & CT & Liver & 131 & Manual & 2 & Tumor \\ 

\cellcolor{HANAASAGI!27} 2017 & ACDC~\cite{ACDCData} & \href{https://www.creatis.insa-lyon.fr/Challenge/acdc/index.html}{\color{green}\faCheckCircle} & MRI & Heart & 150 & Manual & 3 & Organ Parts \\

\cellcolor{HANAASAGI!30} 2018 & FUMPE~\cite{masoudi2018new_FUMPE} & \href{https://www.kaggle.com/datasets/andrewmvd/pulmonary-embolism-in-ct-images}{\color{green}\faCheckCircle} & CT & Lung & 35 & Exp.+Mdl. & 1 & Lesion \\

\cellcolor{HANAASAGI!33} 2018& MSD~\cite{antonelli2022medical} & \href{http://medicaldecathlon.com/}{\color{green}\faCheckCircle} & CT, MRI & Multiple & 1411 CT, 1222 MRI & Manual & 18 & Multi-Task \\

 \cellcolor{HANAASAGI!36} 2018 & DRIVE~\cite{staal2004ridge_DRIVE} & \href{https://drive.grand-challenge.org/DRIVE/}{\color{green}\faCheckCircle} & Fundus & Retina
 & 40 & Manual & 1 & Organ Parts \\

 \cellcolor{HANAASAGI!39} 2018 &  REFUGE~\cite{orlando2020refuge} & \href{https://refuge.grand-challenge.org/}{\color{green}\faCheckCircle} & Fundus & Retina
 & 1200 & Manual & 2 & Organ Parts \\

\cellcolor{HANAASAGI!42} 2019 & CHAOS~\cite{CHAOSdata2019, CHAOS2021, kavur2019} & \href{https://chaos.grand-challenge.org/Combined_Healthy_Abdominal_Organ_Segmentation/}{\color{green}\faCheckCircle} & CT, MRI & Abdomen & 40 CT, 40 MRI & Manual & 4 & Multi-Organ \\

\cellcolor{HANAASAGI!45} 2019 & SIIM-ACR Pneumothorax~\cite{pneumothorax_segmentation} & \href{https://www.kaggle.com/c/siim-acr-pneumothorax-segmentation}{\color{green}\faCheckCircle} & X-Ray & Chest
 & 12047 & Manual & 1 & Lesion \\

 \cellcolor{HANAASAGI!48} 2019 & AbdomenUS~\cite{vitale2020improving} & \href{https://www.kaggle.com/datasets/ignaciorlando/ussimandsegm}{\color{green}\faCheckCircle} & Ultrasound & Abdomen
 & 61 Real, 926 Synth. & Real+Synth. & 8 & Multi-Organ \\

\cellcolor{HANAASAGI!51} 2019 & Breast Ultrasound Images~\cite{al2020dataset} & \href{https://www.kaggle.com/datasets/aryashah2k/breast-ultrasound-images-dataset}{\color{green}\faCheckCircle} & Ultrasound & Breast
 & 780 & Manual & 3 & Tumor \\

 \cellcolor{HANAASAGI!54} 2019 & CAMUS~\cite{leclerc2019deep} & \href{https://www.creatis.insa-lyon.fr/Challenge/camus/}{\color{green}\faCheckCircle} & Ultrasound & Heart
 & 500 & Manual & 3 & Organ Parts \\

\cellcolor{HANAASAGI!57} 2020 & M\&Ms~\cite{MMsData} & \href{https://www.ub.edu/mnms/}{\color{green}\faCheckCircle} & MRI & Heart & 375 & Manual & 3 & Organ Parts \\

\cellcolor{HANAASAGI!60} 2020 & MosMed COVID-19~\cite{morozov2020mosmeddata} & \href{https://www.kaggle.com/datasets/andrewmvd/mosmed-covid19-ct-scans}{\color{green}\faCheckCircle} & CT & Lung & 50 & Manual & 1 & Infection \\

\cellcolor{HANAASAGI!63} 2020 & COVID-19 Radiography~\cite{chowdhury2020can, rahman2021exploring} & \href{https://www.kaggle.com/datasets/tawsifurrahman/covid19-radiography-database?select=COVID-19_Radiography_Dataset}{\color{green}\faCheckCircle} & X-Ray & Chest
 & 21165 & Manual & 1 & Single Organ \\

 \cellcolor{HANAASAGI!66} 2021 & COVID-QU-Ex~\cite{tahir2021covid, COVID-QU-Ex, degerli2021covid, chowdhury2020can, rahman2021exploring} & \href{https://www.kaggle.com/datasets/anasmohammedtahir/covidqu}{\color{green}\faCheckCircle} & X-Ray & Chest
 & 33920 & Manual & 2 & Infection \\

 \cellcolor{HANAASAGI!69} 2021 & QaTa-COV19~\cite{QaTa} & \href{https://www.kaggle.com/datasets/aysendegerli/qatacov19-dataset}{\color{green}\faCheckCircle} & X-Ray & Chest
 & 9258 & Manual & 1 & Infection \\

 \cellcolor{HANAASAGI!72} 2021 & CT2US~\cite{song2022ct2us} & \href{https://www.kaggle.com/datasets/siatsyx/ct2usforkidneyseg}{\color{green}\faCheckCircle} & Ultrasound & Abdomen
 & 4586 & Synth. & 1 & Single Organ \\

 \cellcolor{HANAASAGI!75} 2021 & PolypGen~\cite{ali2021polypgen,ali2022assessing,ali2021deep} & \href{https://www.synapse.org/#!Synapse:syn26376615/wiki/613312}{\color{green}\faCheckCircle} & Endoscope & Colon
 & 8037 & Manual & 1 & Polyp \\

\cellcolor{HANAASAGI!78} 2022 & AbdomenCT-1K~\cite{abdomenct-1k} & \href{https://abdomenct-1k-fully-supervised-learning.grand-challenge.org/}{\color{green}\faCheckCircle} & CT & Abdomen
 & 1112 &  Exp.+Mdl. & 4 & Multi-Organ \\

 \cellcolor{HANAASAGI!81} 2022 & AMOS~\cite{ji2022amos} & \href{https://amos22.grand-challenge.org/}{\color{green}\faCheckCircle} & CT, MRI & Abdomen
 & 500 CT, 100 MRI & Exp.+Mdl. & 15 & Multi-Organ \\
 
\cellcolor{HANAASAGI!84} 2023 & KiTS~\cite{heller2023kits21} & \href{https://kits-challenge.org/kits23/}{\color{green}\faCheckCircle} & CT & Kidney
 & 599 & Exp.+Mdl. & 3 & Organ, Tumor \\

\cellcolor{HANAASAGI!87} 2023 & TotalSegmentator~\cite{Wasserthal_2023_TotalSegmentator} & \href{https://zenodo.org/record/8367088}{\color{green}\faCheckCircle} & CT & Full Body
 & 1228 & Manual & 117 & Multi-Organ \\

\cellcolor{HANAASAGI!90} 2023 & BraTS~\cite{karargyris2023federated, baid2021rsna, menze2014multimodal, bakas10segmentation, bakas2017advancing, bakas2017segmentation, labella2023asnr, moawad2023brain, kazerooni2023brain, adewole2023brain} & \href{https://www.synapse.org/#!Synapse:syn51156910/wiki/622351}{\color{green}\faCheckCircle} & MRI & Brain
 & 4500 & Manual & 3 & Tumor \\

 \cellcolor{HANAASAGI!93} 2023 & HaN-Seg~\cite{podobnik2023han} & \href{https://han-seg2023.grand-challenge.org/}{\color{green}\faCheckCircle} & CT, MRI & Head \& Neck
 & 56 CT, 56 MRI & Manual & 30 & Multi-Organ \\

 \cellcolor{HANAASAGI!96} 2023 & FH-PS-AOP~\cite{FH-PS-AOP} & \href{https://ps-fh-aop-2023.grand-challenge.org/}{\color{green}\faCheckCircle} & Ultrasound & Transperineal
 & 6224 & Exp.+Mdl. & 2 & Multi-Organ \\

\bottomrule[1.5pt]
\end{tabular}
\label{tab:t2i-dataset}
\vspace{-2mm}
\end{table*}

\subsection{Versatile Medical Image Modalities}
Imaging modality is one of the most critical property of a medical image, which has a great impact on developing multi-task models \cite{ma2023segment}. While in general domain ``image'' itself is a modality in comparison with text or audio, the ``modality'' of medical images is defined by the mechanisms they were produced, e.g. CT, MRI, X-Ray, etc.. Because the images of each modality are constructed by sensing different signals (such as X-Ray, ultrasound, magnetic resonance) to reflecting the distribution of tissues in human body, the modalities are significantly different in terms of dimensionality, resolution, noise level, number of channels, etc.. In general, medical imaging modalities can be divided into the following categories.

\begin{itemize}
    \item 3D Images: CT, MRI
    \item 2D Images: 
    \begin{itemize}
        \item Clinical: X-Ray, ultrasound, camera images (endoscope, fundus, dermoscope, etc.)
        \item Pathology
    \end{itemize}
\end{itemize}

\begin{figure*}[h!]
    \centering
    \begin{subfigure}{.39\textwidth}
        \centering
        \includegraphics[trim=170 100 0 0,clip,width=\linewidth]{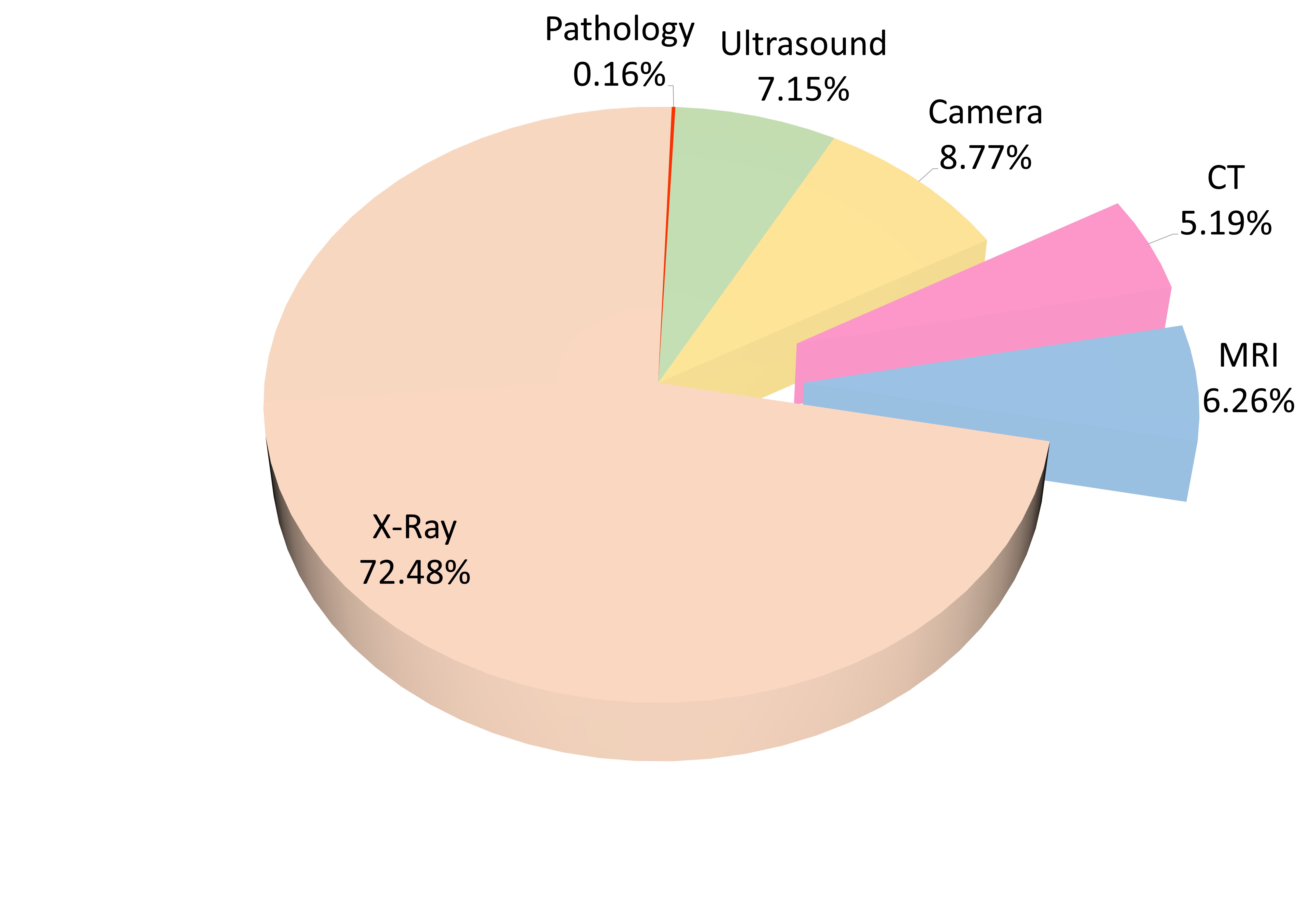}
        \caption{Data distribution by imaging modality}
        \label{fig:modality}
    \end{subfigure}%
    \begin{subfigure}{.61\textwidth}
        \centering
        \includegraphics[width=\linewidth]{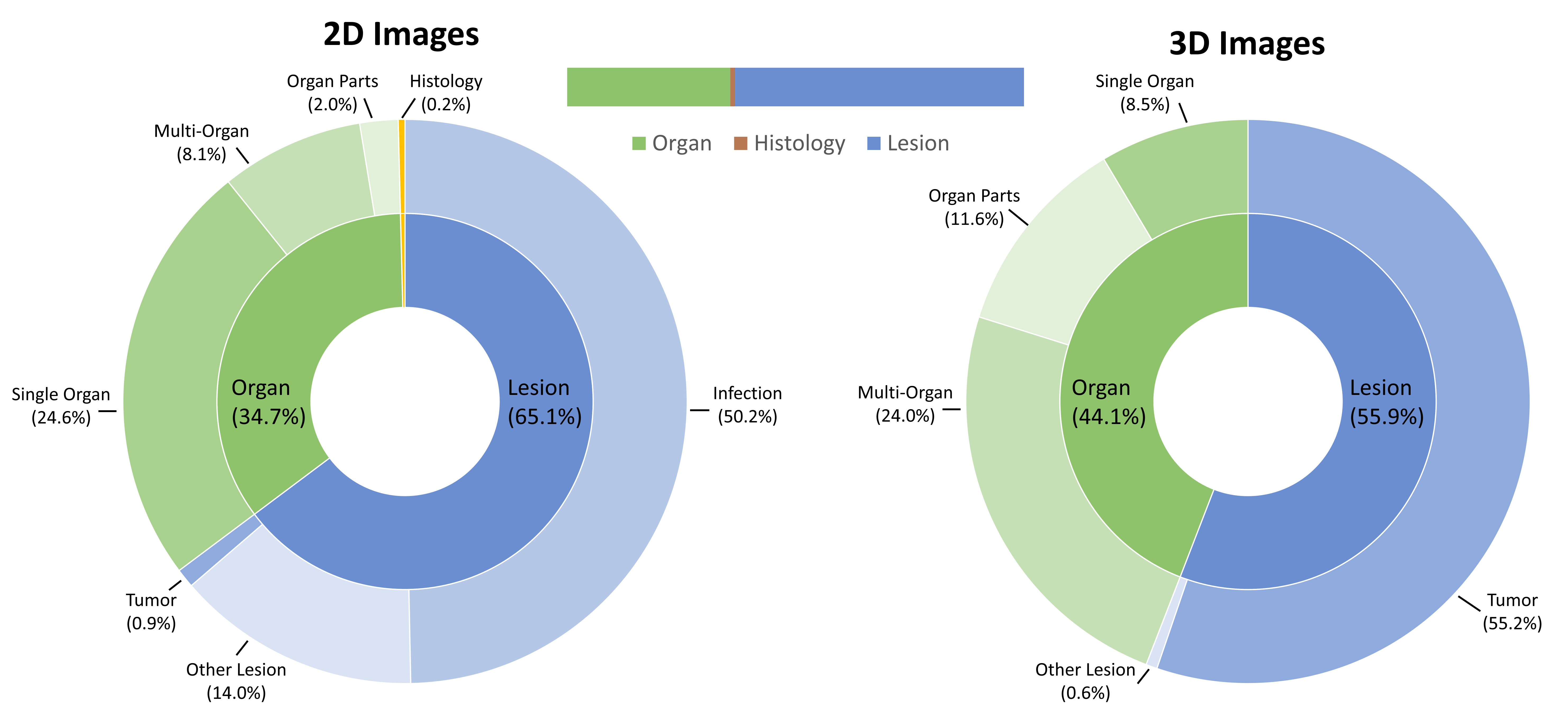}
        \caption{Data distribution by segmentation target}
        \label{fig:target}
    \end{subfigure}
    \caption{\textbf{The distribution of medical image segmentation data} by number of scans/images in the collection of public datasets in Table \ref{tab:t2i-dataset}. Inclusion criteria of the pie charts are: 1. the images are real world data, 2. the annotation process involved domain experts.}
    \label{fig:data_pies}
\end{figure*}

In Figure \ref{fig:modality} we show the distribution of the medical imaging modalities in the collection of public datasets. Among all the modalities, X-Rays composes the largest portion in terms of number of scans. It is worth-noting that over 80\% of the X-Ray data were curated during the COVID-19 pandemic, thanks to the effort by numerous researchers in response to the public health crisis. Ultrasound and camera imaging take similar portion of the 2D images. In contrast, pathology images only take a tiny portion of 0.16\%. Compared to other types of images obtained in clinical practices, pathology images are less in availability and harder for manual annotation, making the development of high accuracy pathological models much more difficult.

The two 3D imaging modalities, CT and MRI, are nearly equal in their percentage. Although the total number of 3D images only takes about one ninth in terms of number of scans, each of them contains a number of 2D slices at the scale of 10-100. When the 3D data are used for models like SAM which take in 2D images, the most common practice is to use the 2D slices for training and evaluation, giving the 3D images much higher weights in developing multi-task models. On the other hand, the 2D slices from the same scan share spatial connection, posing extra challenge to adapt SAM to a higher dimension. The resolution and the spatial distance between the slices also vary across images and datasets (see Table \ref{tab:3d-dataset}) in 3D adaptation.

\subsection{Fine-grained Medical Segmentation Tasks}
While general domain segmentation tasks are usually divided into semantic, instance, or panoptic segmentation, medical image segmentation is more task-tailored \cite{azad2022medical}. Specifically:
\begin{itemize}
    \item Segmentation is usually performed on a fixed vision field (certain part of the human body to perform the scan).
    \item The aim is to extract certain target(s) rather than segmenting the whole vision field.
\end{itemize}

In general, the segmentation targets can be divided as \footnote{Pathology images in principle are different from other types of clinical images with segmentation target at sell level. Their population is also tiny in the datasets, thus requires separate discussion.}:
\begin{itemize}
    \item Organs: the normal tissues parts of the body that are persistent for a certain scan field. The target(s) can be: a single organ, multiple organs in the scan field, and specific parts inside an organ.
    \item Lesion: the abnormality in the body, e.g. tumor, infection, and various types of other lesions.
\end{itemize}

The distribution of segmentation targets in the datasets is shown in Figure \ref{fig:target}, with detailed statistics calculated for 2D and 3D images separately. Lesion data outnumber organ data for both 2D and 3D images, but the fine-grained distribution of the two types of modalities differs a lot. 2D images typically exhibit coarser presentation of the tissues in body, thus are good to segment single large organs like lung, but not less used to show multiple organs or finer structures inside an organ. On the lesion side, the majority of the data are for infection segmentation due to the large number of lung infection data for chest X-Ray. 

On the other hand, 3D images are able to provide a full reconstruction of the spatial structure in the human body, thus enables fine-grained segmentation like multi-organ and organ parts segmentation. On the lesion side, most of the segmentation targets are various types of tumor, as the 3D imaging technologies like CT and MRI are widely used to identify and measure the tumor. In application, having a good estimate of the size of the tumor is critical for cancer diagnosis and staging, demanding precise segmentation tools.

\section{Adaptation of SAM to the Medical Domain}
Following the introduction of SAM on April 5, 2023,  as illustrated in Figure \ref{fig:timeline}, there has been a marked uptrend in its adoption in the medical domain. Through a systematic review and analysis of the existing works, we identify and categorize the adaptation of SAM into four primary methodologies.

\begin{figure*}[h!]
\centering
 \includegraphics[width=\linewidth]{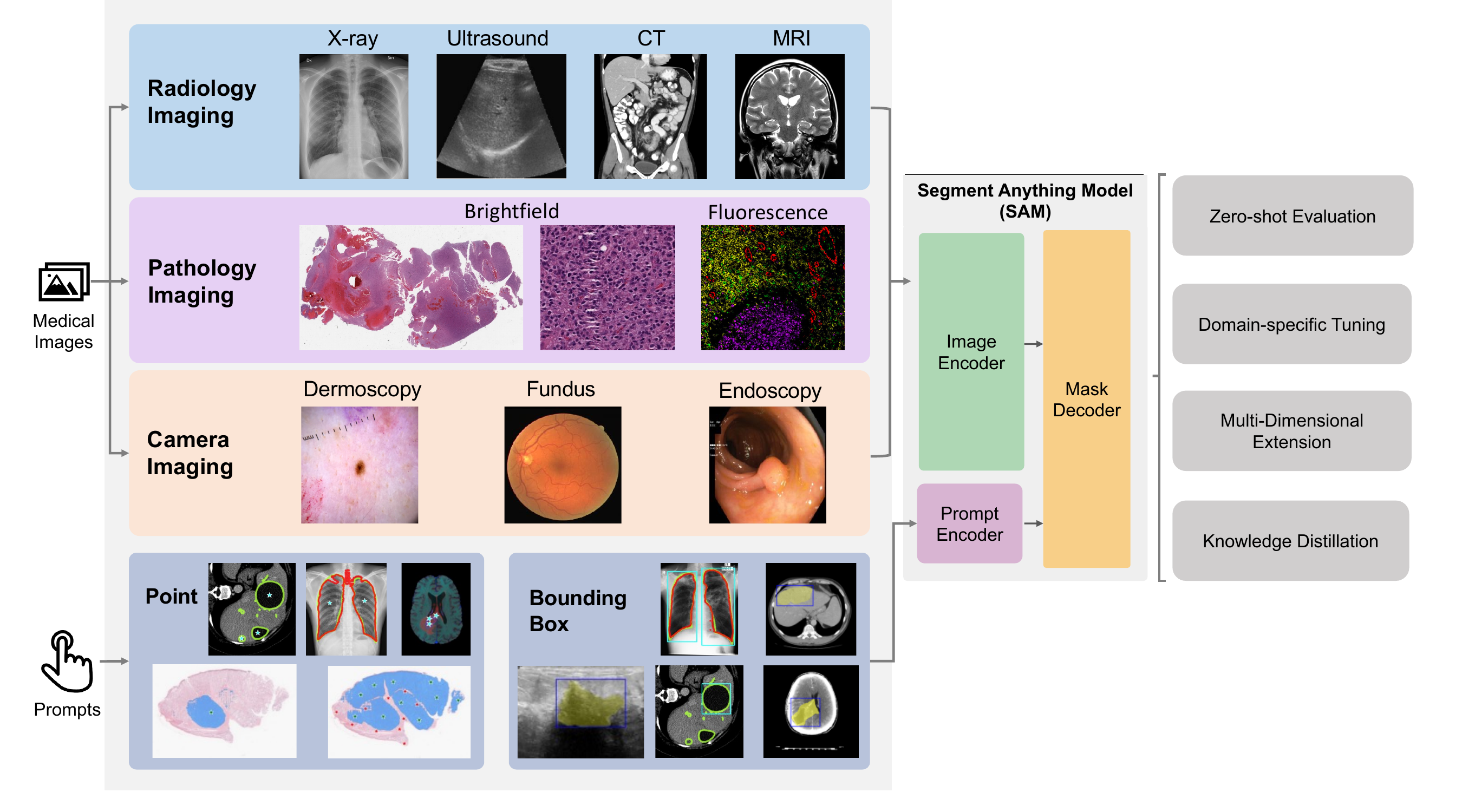}
\label{fig:sam-modalities}
 \caption{\textbf{Application of SAM Across Medical Imaging Modalities.} The figure showcases Radiology, Pathology, and Camera Imaging examples. Central components of SAM, including the Image Encoder, Mask Decoder, and Prompt Encoder, are delineated. Methods ranging from Zero-shot Evaluation to Knowledge Distillation are accentuated within tan boxes.}
\end{figure*}

\begin{figure*}[h!]

\centering
\includegraphics[width=\linewidth]{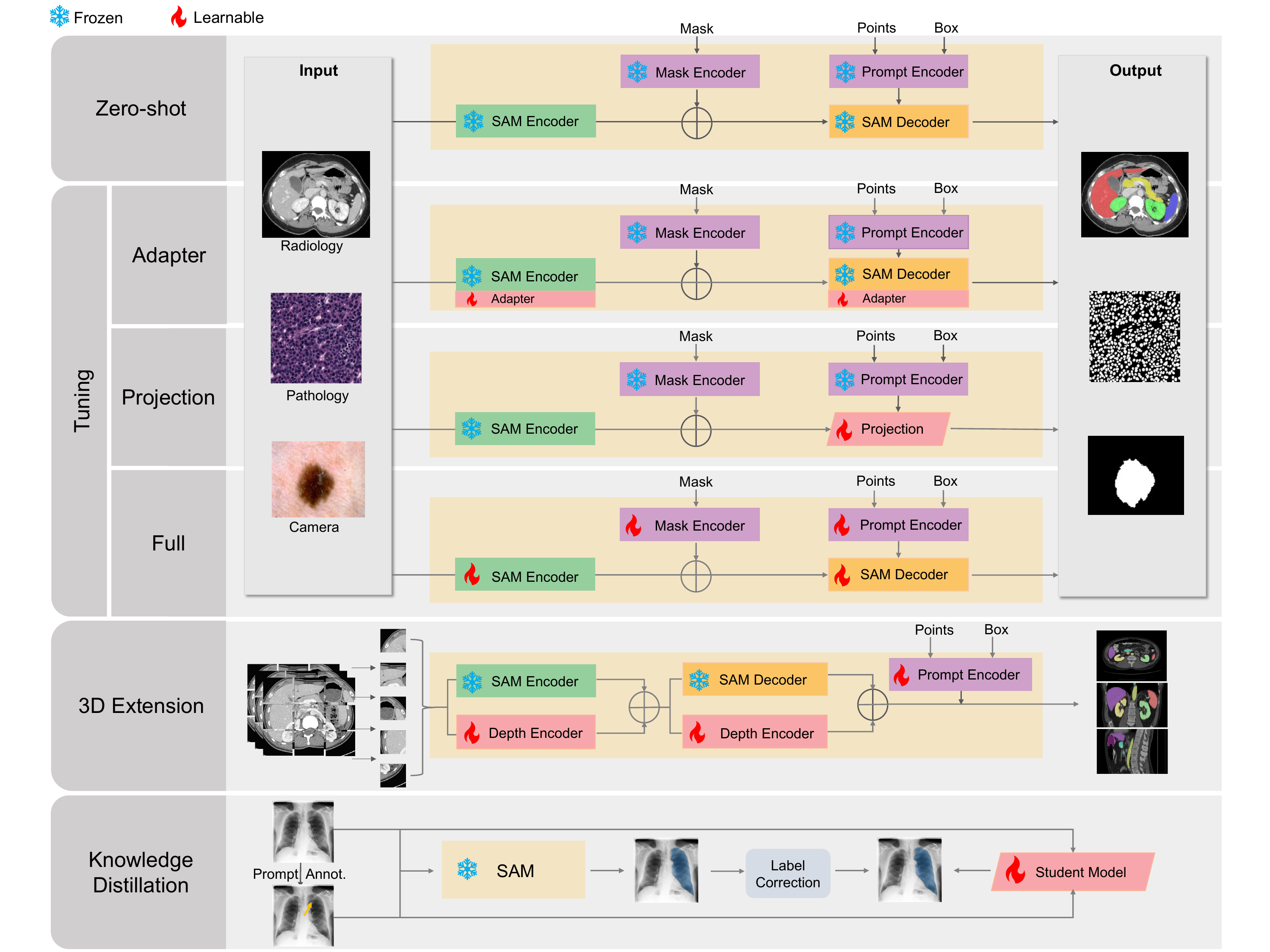}
\caption{\textbf{Decomposition of SAM Adaptation Methods in Medical Imaging.} An illustrative overview of various adaptation strategies of SAM for the medical domain. The figure showcases five key methodologies: \textit{Zero-shot Evaluation}, which assesses SAM's inherent ability for medical image segmentation; \textit{Adapter, Projection, and Full Tuning}, which represent different degrees of model fine-tuning; \textit{3D Extension}, highlighting SAM's adaptation for volumetric data; and \textit{Knowledge Distillation}, where SAM's expertise is transferred to a student model. Each method's flow from input to output, accompanied by specific components and modules, is visualized for clarity.}
\label{fig:sam-arch-decomposistion}
\end{figure*}

\begin{table}[]
\centering
\fontsize{6}{8}\selectfont
\caption{\textbf{A comprehensive list of SAM-Related approaches.} }
\begin{tabular}{c||l|l}
\toprule[1.5pt]
\multicolumn{2}{c} {\textbf{Proposed Perspectives}}  & \multicolumn{1}{c}{\textbf{Years: Methods}}\\
\midrule
\multirow{12}{*}{\makecell*[c]{Zero-Shot}} & Robustness Evaluation &{\makecell*[l]{\textbf{2023 April:}~\cite{deng2023segment},~\cite{roy2023sam},~\cite{zhou2023can}, \\ \cite{hu2023sam}, \cite{he2023accuracy}, \cite{mazurowski2023segment}, \cite{shi2023generalist},  ~\cite{ji2023segment},  \\ ~\cite{wang2023sam}, ~\cite{cheng2023sam}, \cite{ji2023sam}, \\\textbf{2023 May:}~\cite{mattjie2023exploring}, ~\cite{hu2023breastsam}, ~\cite{iytha2023lung}\\\textbf{2023 June:}~\cite{ahmadi2023comparative}, ~\cite{hu2023efficiently}, ~\cite{blumenstiel2023mess} \\\textbf{2023 July:}~\cite{chen2023ability} \\\textbf{2023 August:}~\cite{wang2023sam}, \\\textbf{2023 September:} ~\cite{ranem2023exploring}}}  \\  

& Image / Prompt Augmentation &{\makecell*[l]{\textbf{2023 April:}~\cite{zhang2023input}, \cite{yan2023piclick}\\\textbf{2023 June:}~\cite{shen2023temporally}, ~\cite{lei2023medlsam} \\\textbf{2023 July:}~\cite{deng2023sam}, ~\cite{dai2023samaug} \\\textbf{2023 August:}~\cite{biswas2023polyp}, ~\cite{wu2023selfprompting}, ~\cite{yao2023false}}}  \\  

& Clinical Applications &{\makecell*[l]{\textbf{2023 April:}~\cite{liu2023samm}, ~\cite{wang2023gazesam}, \cite{putz2023segment}, \cite{mohapatra2023brain} \\\textbf{2023 May:}~\cite{lee2023iamsam}, ~\cite{sun2023pathasst} \\\textbf{2023 June:}~\cite{zhang2023segment}, ~\cite{semeraro2023tomosam} \\\textbf{2023 July}: ~\cite{cheng2023axoncallosumem}, ~\cite{hossain2023robust}, ~\cite{li2023auto}}}  \\
\midrule
\multirow{12}{*}{\makecell*[c]{ Tuning }}
& Adapter &{\makecell*[l]{\textbf{2023 April:}~\cite{chen2023sam}, \cite{zhang2023customized}, \\\textbf{2023 June:}~\cite{gao2023desam}, ~\cite{wang2023mathrm} \\\textbf{2023 August:} ~\cite{shin2023cemb}}}\\


& Projection &{\makecell*[l]{\textbf{2023 April:}~\cite{zhang2023customized} \\\textbf{2023 June:}~\cite{hu2023efficiently}, ~\cite{horst2023cellvit} \\\textbf{2023 July:}~\cite{cui2023all}, ~\cite{zhang2023sam}, ~\cite{shi2023cross} \\\textbf{2023 August}~\cite{feng2023cheap}, ~\cite{zhang2023self}}} \\

 & Full &{\makecell*[l]{\textbf{2023 April:}~\cite{hu2023skinsam} \\\textbf{2023 May:}~\cite{li2023polyp} \\\textbf{2023 June:}~\cite{horst2023cellvit} \\\textbf{2023 July:}~\cite{shi2023cross} \\\textbf{2023 August:}~\cite{paranjape2023adaptivesam}}}  \\  

\midrule
\makecell{Multi-Dimensional\\Adaptation}

& 3D Extension &{\makecell*[l]{\textbf{2023 June:}~\cite{gong20233dsam} \\\textbf{2023 September:} ~\cite{bui2023sam3d}, ~\cite{chen2023ma}, ~\cite{kim2023medivista}}}  \\  
\midrule
\makecell{Knowledge\\Distillation}
& {\makecell*[l]{Weakly/Semi-Supervised \\ Learning with Pseudo Prior}}  &{\makecell*[l]{\textbf{2023 May:}~\cite{he2023weakly}, ~\cite{li2023samscore}\\\textbf{2023 June:}~\cite{kellener2023utilizing}, ~\cite{li2023segment} \\\textbf{2023 July:}~\cite{cui2023all}, ~\cite{zhang2023sam} \\\textbf{2023 August:}~\cite{huang2023push}, ~\cite{zhang2023samdsk}}}  \\  
\bottomrule[1.5pt]
\end{tabular}
\label{tab:T2I}
\vspace{-2mm}
\end{table}

\begin{table*}[]
\centering
\fontsize{7}{9.6}\selectfont
\caption{\textbf{Overview of various existing works that build upon SAM.}}
\begin{tabular}{cl||cc|cc|cc|ccccc}
\toprule[1.5pt]
\small \textbf{Year-Month} & \diagbox {\small \textbf{Method}}{\small \textbf{Tasks}} & 2D & 3D & A.P.P & P.A & E.Frozen & E.Finetune & R.N.M & T.P.H & T.P.E & T.A & Downstream Tasks\\
\midrule
\cellcolor{HIWA!45} 2023-April & SAM-Adaptor~\cite{chen2023sam} & \usym{2714} & - & - & - & \usym{2714} & - & - & - & - & \usym{2714} & Polyp \\
\cellcolor{HIWA!45} 2023-April & SAMAug~\cite{zhang2023input} & \usym{2714} & - & - & \usym{2714} & \usym{2714} & - & \usym{2714} & - & - & - & H\&E, Polyp\\
\cellcolor{HIWA!45} 2023-April & MedSAM Adaptor~\cite{wu2023medical} & \usym{2714} & \usym{2714} & - & - & - & - & \usym{2714} & - & - & \usym{2714} & Abd, Opt, B.T, T.N\\
\cellcolor{HIWA!45} 2023-April & LOSAM~\cite{zhang2023input} & \usym{2714} & - & - & - & \usym{2714} & - & - & - & \usym{2714} & - & Vessel \& Lesion \\
\cellcolor{HIWA!45} 2023-April & SAMed~\cite{zhang2023customized} & \usym{2714} & - & - & - & \usym{2714} & - & - & \usym{2714} & \usym{2714} & \usym{2714} & Abd \\
\cellcolor{HIWA!45} 2023-April & GazeSAM~\cite{wang2023gazesam} &  \usym{2714} & - & \usym{2714} & - & \usym{2714} & - & - & - & - & - & Abd \\
\cellcolor{HIWA!45} 2023-April & SkinSAM~\cite{hu2023skinsam} & \usym{2714} & - & - &  \usym{2714} & - & \usym{2714} & - & - & - & - & S.L \\
\cellcolor{HIWA!45} 2023-April & PiClick~\cite{yan2023piclick} & \usym{2714} & - & - &  - & \usym{2714} & - & - & - & - & - & Neural Tissue\\
\cellcolor{HIWA!65} 2023-May & Polyp-SAM~\cite{li2023polyp} & \usym{2714} & - & - &  - & - & \usym{2714} & - & - & \usym{2714} & - & Polyp\\
\cellcolor{HIWA!65} 2023-May & SAM-Track~\cite{cheng2023segment} & \usym{2714} & - & - &   \usym{2714} &  \usym{2714} & - & - & - & - & - & Brain\\
\cellcolor{HIWA!65} 2023-May & WS-SAM~\cite{he2023weakly} & \usym{2714} & - & \usym{2714} & - & \usym{2714} & - & \usym{2714} & - & - & - & Polyp\\
\cellcolor{HIWA!65} 2023-May & BreastSAM~\cite{hu2023breastsam} & \usym{2714} & - & - & \usym{2714} & \usym{2714} & - & - & - & - & - & Breast C.\\
\cellcolor{HIWA!65} 2023-May & LuSAM~\cite{iytha2023lung} & \usym{2714} & - & - & - & \usym{2714} & - & - & - & - & - & Lung\\
\cellcolor{HIWA!65} 2023-May & IAMSAM~\cite{lee2023iamsam} & \usym{2714} & - & - & - & \usym{2714} & - & - & - & - & - & H \& E\\
\cellcolor{HIWA!75} 2023-June & DeSAM~\cite{gao2023desam} & \usym{2714} & - & - & - & \usym{2714} & - & - & - & \usym{2714} & \usym{2714} & Prostate\\
\cellcolor{HIWA!75} 2023-June & AutoSAM(1)~\cite{shaharabany2023autosam} & \usym{2714} & - & - & - & \usym{2714} & - & - & - & \usym{2714} & - & H \& E, Polyp\\
\cellcolor{HIWA!75} 2023-June & TEPO~\cite{shen2023temporally} & \usym{2714} & - & \usym{2714} & - & \usym{2714} & - & \usym{2714} & - & - & - & Brain \\
\cellcolor{HIWA!75} 2023-June & RASAM~\cite{zhang2023segment} & \usym{2714} & - & - & - &  \usym{2714} & - & - & - & - & - & Organ-at-risk \\
\cellcolor{HIWA!75} 2023-June & 3DSAM-adaptor~\cite{gong20233dsam} & - & \usym{2714} & - & \usym{2714} &  \usym{2714} & - & - & - & \usym{2714} & \usym{2714} & Parts Tumor \\
\cellcolor{HIWA!75} 2023-June & AutoSAM(2)~\cite{hu2023efficiently} & \usym{2714} & - & - & - &  \usym{2714} & - & - & \usym{2714} & - & - & Cardiac Structure \\
\cellcolor{HIWA!75} 2023-June & MedLSAM~\cite{lei2023medlsam} & \usym{2714} & - & \usym{2714} & - &  \usym{2714} & - & \usym{2714}& - & - & - & H \& N, Abd, Lung \\
\cellcolor{HIWA!75} 2023-June & CellViT~\cite{horst2023cellvit} & \usym{2714} & - & - & - & - & \usym{2714} & - & \usym{2714} & - & - & H \& E \\
\cellcolor{HIWA!85} 2023-July & SAM-U~\cite{deng2023sam} & \usym{2714} & - & - & \usym{2714} & \usym{2714} & - & - & - & - & - & Opt \\
\cellcolor{HIWA!85} 2023-July & SAM$^{\text{Med}}$~\cite{wang2023mathrm} & \usym{2714} & - & \usym{2714} & \usym{2714} & \usym{2714} & - & - & - & - & \usym{2714} & Abd, Prostate \\
\cellcolor{HIWA!85} 2023-July & SAMAug~\cite{dai2023samaug} & \usym{2714} & - & - & \usym{2714} & \usym{2714} & - & - & - & - & - & Polyp, Lung \\
\cellcolor{HIWA!85} 2023-July & All-in-SAM~\cite{cui2023all} & \usym{2714} & - & \usym{2714} & \usym{2714} & \usym{2714} & - & - & \usym{2714} & \usym{2714} & - & H \& E \\
\cellcolor{HIWA!85} 2023-July & SAM-Path~\cite{zhang2023sam} & \usym{2714} & - & - & \usym{2714} & \usym{2714} & - & - & \usym{2714} & \usym{2714} & - & H \& E \\
\cellcolor{HIWA!85} 2023-July & CmAA~\cite{shi2023cross} & \usym{2714} & - & - & - & \usym{2714} & - & - & \usym{2714} & - & - & Glioma \\
\cellcolor{HIWA!85} 2023-July & MedSAM~\cite{ma2023segment} & \usym{2714} & - & - & - & - & - & \usym{2714} & - & - & - & 15 I.M, >30 C.T \\
\cellcolor{HIWA!90} 2023-August & SAM-MLC~\cite{huang2023push} & \usym{2714} & - & \usym{2714} & - & \usym{2714} & - & \usym{2714} & - & - & - & Lung \\
\cellcolor{HIWA!90} 2023-August & AdaptiveSAM~\cite{paranjape2023adaptivesam} & \usym{2714} & - & - & \usym{2714} & - & \usym{2714} & - & \usym{2714} & \usym{2714} & - & S.S \\
\cellcolor{HIWA!90} 2023-August & Poly-SAM++~\cite{biswas2023polyp} & \usym{2714} & - & - & \usym{2714} & \usym{2714} & - & - & - & - & - & Polyp \\
\cellcolor{HIWA!90} 2023-August & SPSAM~\cite{wu2023selfprompting} & \usym{2714} & - & \usym{2714} & \usym{2714} & \usym{2714} & - & - & - & \usym{2714} & - & Polyp, S.L \\
\cellcolor{HIWA!90} 2023-August & SamDSK~\cite{zhang2023samdsk} & \usym{2714} & - & - & - & \usym{2714} & - & \usym{2714} & - & - & - & Polyp, S.L, Breast C. \\
\cellcolor{HIWA!90} 2023-August & AutoSAM Adaptor~\cite{li2023auto} & - & \usym{2714} & - & \usym{2714} & \usym{2714} & - & - & - & \usym{2714} & \usym{2714} & Abd \\
\cellcolor{HIWA!90} 2023-August & SAM-Med2D~\cite{cheng2023sam} & \usym{2714} & - & - & - & \usym{2714} & - & - & - & \usym{2714} & \usym{2714} & 9 MICCAI2023 \\
\cellcolor{HIWA!90} 2023-August & SAMedOCT~\cite{fazekas2023samedoct} & \usym{2714} & - & - & - & \usym{2714} & - & \usym{2714} & - & - & - & OCT \\
\cellcolor{HIWA!90} 2023-September & SAM3D~\cite{bui2023sam3d} & \usym{2714} & - & - & - & \usym{2714} & - & - & \usym{2714} & - & - & Brain Lung,, Abd \\
\cellcolor{HIWA!90} 2023-September & SAMUS~\cite{lin2023samus} & \usym{2714} & - & - & - & \usym{2714} & - & \usym{2714} & - & - & - & Ultrasound \\
\cellcolor{HIWA!90} 2023-September & MA-SAM~\cite{chen2023ma} & - & \usym{2714} & - & - & \usym{2714} & - & - & - & - & \usym{2714} & Abd, Prostate, S.S \\
\cellcolor{HIWA!90} 2023-September & MedVISTA-SAM~\cite{chen2023ma} & \usym{2714} & \usym{2714} & - & - & \usym{2714} & - & - & - & - & \usym{2714} & Echocardiography \\

\bottomrule[1.5pt]
\end{tabular}
\vspace{0.5mm}
\raggedright
\footnotesize{{\underline{<\textbf{Acronym}: Meaning>}} \textbf{A.P.P}:Adapt Psuedo Prior; \textbf{PA}:Prompt Augmentation; \textbf{E.}:Encoder; \textbf{R.N.M}: Retrain New Model; \textbf{T.P.H}:Train Projection Head; \textbf{T.P.E}:Train Prompt Encoder; \textbf{T.A}:Train Adaptor;  \textbf{Abd}:Abdomen; \textbf{Opt}:Optic; \textbf{B.T}:Brain Tumor; \textbf{T.N}:Thyroid Nodule; \textbf{I.M}: Imaging Modalities; \textbf{C.T}: Cancer Types; \textbf{Per.}: Peripheral; \textbf{C.}: Cancer 
}
\label{tab:multi_tasks}
\vspace{-2mm}
\end{table*}

\subsection{Zero-shot Segmentation Capabilities Evaluation}
The zero-shot evaluation illuminates SAM's ability to adapt to medical imaging tasks without any prior domain-specific training \cite{kirillov2023segment}. Such zero-shot capabilities primarily assesses whether SAM, known for its proficiency in general domain image processing, can effectively generalize its learned patterns to the nuanced and variable-rich world of medical imaging.

Medical imaging presents unique challenges, distinguished by factors like varied imaging protocols and a wide range of patient demographics \cite{deng2023segment, roy2023sam, zhou2023can, ji2023sam, hu2023breastsam}. These complexities are not as predominant in standard domain images, making SAM's adaptability in this context particularly intriguing. Comprehensive quantitative analyses have been conducted to compare SAM's zeroshot results with current SOTA techniques in various target areas and regions. For detailed performance metrics and observations, readers are referred to Section 5.1.

\subsection{Domain-Specific Tuning}

The versatility of SAM in instance segmentation, while noteworthy, has demonstrated varying results across different contrast appearances and organ morphologies inherent to medical images. To address this, researchers have explored several domain-specific tuning strategies. This section provides an objective overview of three prominent strategies: 
\subsubsection{Projection Tuning} One approach seeks to capitalize on the rich knowledge obtained from natural image domains. By replacing the pretrained decoder with a new, task-specific projection head, the model aims to harness generalized features (e.g., edges, boundaries) that are prevalent in datasets like ImageNet. As shown in Figure \ref{fig:sam-arch-decomposistion}, recent works propose to replace the projection head into various architectures and train it from scratch, such as multi-layer perceptrons (MLPs), convolutions, or vision transformers \cite{zhang2023customized, hu2023efficiently, feng2023cheap}. While this strategy aims to maintain knowledge from the natural image domain, its effectiveness can be contingent upon the quality of the prompt, especially when dealing with significant domain shifts inherent to multi-modal images \cite{deng2023segment, cui2023all}.
\subsubsection{Adapter Tuning} To bolster the adaptability of SAM across various imaging modalities, recent research has focused on incorporating adapter modules within the SAM encoder blocks, as illustrated in Figure \ref{fig:sam-arch-decomposistion}. These adapters, integrated without modifying the core architecture, are designed to fine-tune the model's response to the specific challenges presented by medical imaging \cite{chen2023sam, gao2023desam, wang2023mathrm, shin2023cemb}. This arrangement facilitates a blend of pre-existing knowledge from natural images with nuanced, task-specific insights from the medical domain. Yet, as indicated in Table \ref{tab:T2I}, there have been relatively few studies that have developed these adapter modules from the ground up specifically for domain adaptation \cite{shi2023cross}. Consequently, the optimal configuration, placement, and overall impact of these adapters within the SAM framework continue to be subjects of active exploration in the field.
\subsubsection{Full Tuning} Refining the entire SAM architecture to more closely resonate with the subtleties of medical imaging represents a comprehensive strategy. This approach entails not just minor adjustments but a substantial reconfiguration, fine-tuning both the encoder and decoder of SAM to transition its generalized knowledge base, originally derived from natural images, to a context more specific to the medical domain. While this method promises a deeper integration with the unique characteristics of medical data, it comes with the caveat of potentially requiring more intensive training efforts and greater resource allocation. Moreover, recent studies have highlighted the difficulties in fine-tuning SAM's architecture, especially when dealing with a limited dataset \cite{}. Thus, identifying an optimal approach to adapt and leverage SAM’s foundational structure for medical applications, in a manner that efficiently incorporates medical knowledge, remains a key challenge.

\subsection{3D Imaging Modalities Extension}
In the field of medical imaging, most protocols generate imaging modalities in three-dimensional (3D) format. To align with SAM’s two-dimensional (2D) framework, axial slices are typically processed to fit a 2D context, with predictions made on a slice-by-slice basis for 3D images. This approach, however, does not fully account for the spatial relationships between slices, often resulting in imprecise organ boundary alignments. To mitigate this, pioneering efforts have adopted a 2.5D approach, incorporating additional convolutional modules that compute through-plane features to enhance the spatial correlation between slices within the SAM encoder blocks \cite{wu2023medical, gong20233dsam}. Yet, this method still focuses on axial and through-plane spatial correlations on a slice-by-slice basis rather than establishing a comprehensive volumetric correspondence. An alternative strategy involves the amalgamation of all slice representations into a single volumetric map, followed by the adaptation of spatial correlations at the feature level using a 3D decoder network \cite{bui2023sam3d}. Despite this advancement, the encoder network continues to depend on the static weights from SAM’s original encoder, which limits robustness when compared to various state-of-the-art transformer networks.


\subsection{Knowledge Distillation}
Beyond the intricacies of model architecture, leveraging SAM’s predictive output as a form of prior knowledge has become an emerging trend in medical imaging. Recent studies typically begin by generating a rough segmentation mask using SAM, guided by carefully crafted prompts. However, using these coarse masks as pseudo labels presents a challenge when training a more accurate 'student' model. To address this, an auxiliary label refinement network is often employed to improve the quality of these initial segmentation labels \cite{huang2023push, cui2023all}. The enhanced labels can then be used to train a task-specific model from the ground up. Moreover, SAM's zero-shot learning capabilities markedly increase the efficiency of generating pseudo labels from vast quantities of unlabeled data. Semi-supervised methodologies incorporating SAM are also being explored to bolster the robustness of task-specific models \cite{zhang2023samdsk}. As such, in the realm of weakly and semi-supervised learning, the distillation of pixel-level insights has become a crucial focus, enabling the efficient utilization of large-scale datasets in model training.

\section{SAM's Performance on Medical Imaging Segmentation}
In parallel, it is also equally crucial to explore its adaptive performance across different anatomical regions and scales. Both the research community and practicing clinicians are particularly keen on ensuring consistent and reliable qualitative performance, especially when evaluating different anatomical regions or adapting different imaging modalities. The distinctive benefit of the concept "zero-shot segmentation" has been pivotal to propel more intensive research, aiming to juxtapose the qualitative performance of current SOTA approaches against the pre-trained SAM. Table \ref{tab:rad}, \ref{tab:path} and \ref{tab:cam} present a bird's-eye view of these qualitative assessments across a spectrum of imaging modalities. To dissect and understand these patterns more comprehensively, we have organized the evaluation into two distinct macro-to-micro aspects: 1) imaging modalities \& downstream tasks and 2) methodologies with SAM's foundation.


\subsection{Macro Aspect: Imaging Modalities \& Downstream Tasks}
\textbf{Radiology Imaging:} We begin our investigation with 2D radiology images, categorizing them into two main modalities: OCT and X-ray. Figure \ref{fig:data-piechart} reveals that X-ray images constitute a significant portion of the data distribution assessed by SAM. However, this modality primarily targets the chest area, delving deeper into specific lung conditions such as lung opacity, viral pneumonia, and tuberculosis, as detailed in Table \ref{tab:rad}. Conversely, there are limited studies showcasing SAM's effectiveness with OCT imaging. When comparing this to 3D radiology imaging, we find it intriguing that both CT and MR modalities encompass nearly the entirety of anatomical regions, contrasting the specific focuses of X-ray and OCT. Moreover, specific anatomical regions, including the brain, head \& neck, and abdomen, are frequently targeted. Given the diverse imaging protocols, distinct boundaries are evident between different tissues and organs. As a result, SAM demonstrates commendable performance across various organs of interest, adapting well to a range of prompt guidance, such as 1 box, 1 p.f, and 2 p.b.

\textbf{Pathology Imaging:} From Table \ref{tab:path}, we observe that initial studies for pathology imaging have been focused on H \& E modality across variable anatomical regions, such as skin, kidney and colon. Comparing to radiology imaging, fine-grained semantics in multi-scale resolution such as cells, nuclei and membranes, are targeted in H \& E imaging. However, with the substantial sparsity between neighboring cells, prompt such as a large bounding boxes are difficult to provide fine-grained guidance and localize the target regions for segmentation. Therefore, we can observe that numerous points as the prompt, demonstrate to have the best performance and generate a fair qualitative performance in different scenarios. However, still a substantial gap in qualitative performances between SAM and current SOTAs are demonstrated in most of the target cases.

\textbf{Camera Imaging:} As camera images are reconstructed in 2D, similar to the format in natural images, SAM demonstrates a good transferrable capability to segment targeted regions with clear boundaries. Most of the semantic targets in Table \ref{tab:cam} are easy to localize within the images and demonstrate a fair quantitative performance such as surgical instruments (Dice: 0.893) and polyp (Dice: 0.872). However, we observe that SAM demonstrates limited robustness to segment fundus vasculature in retinal fundus images. The point prompt is limited to provide an extensive guidance to localize the continuity of the semantic structure, similar to the optic nerve segmentation in head \& neck MR images.


\subsection{Micro Aspect: Methodologies with SAM's Foundation}
With Table \ref{tab:rad}, \ref{tab:path} \& \ref{tab:cam}, we already have a general view on the zero-shot performance of SAM with different imaging modalities and fine-grained tasks. However, we can observe that the quantitative robustness is still limited with the general knowledge prior from natural image domain and the specific prompts guidance. Therefore, researchers begin to provide an alternative that adapts each independent module (e.g., encoder network, mask encoder, prompt encoder) as the foundation architecture and fine-tune for better performance. Table \ref{tab:multi_tasks} demonstrates an overview of variable model architectures with SAM's module as the foundation basis from April 2023 to August 2023. We further summarize our observation in Table 4 as four folds: 

\indent\textbf{Image Dimensionality:} The majority of previous studies have processed 16-bit volumetric images (such as CT and MRI scans) into 8-bit 2D slices for downstream segmentation tasks. However, a minority have begun to adapt SAM to handle 3D volumetric data. Proposals include improving spatial correspondence between slices by adding modules to the encoder and decoder networks that can extract features from through-plane axes (such as the x-z and y-z planes, see \cite{wu2023medical, gong20233dsam}). These extracted features are then either combined with axial features or used to create a 3-dimensional volume through depthwise convolution for feature integration. In addition to module adaptation, another emerging approach is to concatenate slice-wise embeddings into a volumetric representation and integrate this with the decoder network to maintain spatial coherence in segmentation tasks. Given the inherently three-dimensional nature of radiological images, there is a clear demand for direct segmentation of 3D structures. Extending SAM to accommodate 3D inputs shows promising potential for real-world clinical application.


\textbf{Data/Prompt Augmentation:} Another perspective to enhance SAM's performance is the adaptation of prompts and the generated segmentation prior from SAM. Instead of following the traditional human-in-the-loop prompts, multiple works are proposed to generate multiple pseudo prompts (e.g., point prediction \cite{dai2023samaug}, bounding box prediction \cite{lei2023medlsam}) with deep neural networks and apply it to SAM as an end-to-end framework. Particularly for pathology images, numerous cells exists within a gigapixel images and it is difficult to localize multiple cellular regions with a single point / bounding box prompt only. Therefore, multi-instance learning is initially performed to provide a coarse localization on different cells. Such instance segmentation label are then converted into circle-like bounding boxes for prompting and demonstrate a significant enhancement in cell segmentation performance \cite{cui2023all}. Another augmentation technique is to additionally learn the outputs from pretrained SAM. While the segmentation quality is fair, multiple works are proposed to train a label refinement model with the coarse SAM output and leverage the refined output for additional training \cite{cui2023all, huang2023push}. With the introduction of label refinement idea, weakly-supervised learning may be another potential alternative to train a model from scratch with good quality pseudo labels. 

\indent\textbf{Encoder w/o Finetuning:} Beyond the scope of data preparation, considerable efforts have been devoted to developing innovative training strategies using SAM as the foundational model for downstream segmentation tasks. While most recent research has focused on utilizing the pretrained, frozen encoder weights of SAM to accommodate medical images, only a handful of studies have explored the full potential of fine-tuning the encoder network. Given that the encoder has been extensively trained on natural images, acquiring a vast repository of prior knowledge, the prevailing wisdom is to retain this knowledge and enrich it with domain-specific insights from medical images to boost performance. Nevertheless, the peculiarities of different imaging domains present a unique set of challenges, making the establishment of a universally optimal fine-tuning approach elusive. This dilemma extends to the decision of whether to fine-tune the entirety of the encoder network or to selectively fine-tune specific intermodular components within the encoder block. 

\indent\textbf{Additional Training:} A considerable volume of recent research has been directed at integrating SAM with a frozen encoder network, where additional modules are introduced to either substitute specific components (such as the decoder network or prompt encoder) or to implement an adaptor for translating representations between encoder layers. Illustrated in Figure 4, there is a discernible uptick in the number of studies employing freshly trained projection head modules between June and August 2023. In contrast, there is a noticeable decline in strategies that focus on training adaptor modules on a per-layer basis. Notably, in the realm of pathology imaging, there is a predilection towards retraining both the prompt encoder and projection network. Meanwhile, for radiology imaging, the preference shifts to training adaptor modules that operate on distinct blocks of layers. Although traditional prompts have proven to be efficient for organ segmentation tasks, they falter when tasked with adapting to multi-scale semantic segmentation (such as cells and tissues). Consequently, the initial trends differ markedly between imaging modalities.



\begin{table*}[]
\centering
\fontsize{7}{9.6}\selectfont
\caption{\textbf{Quantitative Comparisons between the Zero-Shot Performance of SAM and Current State-Of-The-Art (SOTA) Approaches on Different Radiology Datasets. (p.f: foreground point, p.b: background point)}}
\begin{tabular}{cccccccc}
\toprule[1.5pt]
\multirow{2}{*}{\small Dim.} & \multirow{2}{*}{\small Modality} & \multirow{2}{*}{\small Region} & \multirow{2}{*}{\small Targets} & \multicolumn{3}{c}{\small Performance} & \multirow{2}{*}{\small Prompt Mode}\\
& & & & \small SOTAs & \small MedSAM & \small SAM & \\
\midrule
\multirow{46}{*}{\small 3D} & \multirow{14}{*}{\makecell*[c]{\small CT}} & \multirow{2}{*}{\small Brain} & Intracranial Hemorrhage & 0.795 \cite{li2023state} & 0.940 & 0.867 \cite{ma2023segment} & 1 p.f, 2 p.b  \\
& & & Glioblastoma & 0.913 \cite{hatamizadeh2022swin} & 0.943 & 0.744 \cite{ma2023segment} & 1 p.f, 2 p.b\\
& & \multirow{3}{*}{\small Head \& Neck} & Head-Neck Cancer & 0.788\cite{andrearczyk2021overview} & 0.794 & 0.614 \cite{ma2023segment} & 1 p.f, 2 p.b \\
& & & Lymph Nodes & 0.742 \cite{guo2022thoracic} & 0.821 & 0.771 \cite{ma2023segment} & 1 p.f, 2 p.b \\
& & & Throat Cancer & 0.667 \cite{bai2021deep} & 0.803 & 0.281 \cite{ma2023segment} & 1 p.f, 2 p.b \\

& & \multirow{9}{*}{\makecell*[c]{\small Abdomen}} & Pancreas \& Tumor  & 0.828, 0.623 \cite{liu2023clip} & 0.872, 0.791 & 0.731, 0.741 \cite{ma2023segment} & 1 p.f, 2 p.b \\
& & & Liver \& Tumor & 0.950, 0.790 \cite{liu2023clip} & 0.980, 0.887 & 0.916, 0.766 \cite{ma2023segment} & 1 p.f, 2 p.b \\
& & & Spleen  & 0.974 \cite{isensee2021nnu} & 0.976 & 0.938 \cite{huang2023segment} & 1 box \\
& & & Kidney \& Tumor & 0.948, 0.763 \cite{lee2023deformux} & 0.971, 0.902 & 0.947, 0.867 \cite{ma2023segment} & 1 p.f, 2 p.b \\
& & & Aorta & 0.956 \cite{lee2023scaling} & 0.956 &  0.912 \cite{ma2023segment} & 1 p.f, 2 p.b \\
& & & Esophagus & 0.861 \cite{lee2023scaling} & 0.737 & 0.845 \cite{he2023accuracy} & 1 box \\
& & & Stomach & 0.921 \cite{lee2023scaling} & 0.962 & 0.855 \cite{ma2023segment} & 1 p.f, 2 p.b \\
& & & Gallblader & 0.921 \cite{lee2023scaling} & 0.918 & 0.872 \cite{he2023accuracy} & 1 box \\
& & & IVC & 0.924 \cite{lee2023scaling} & 0.918 & 0.897 \cite{he2023accuracy} & 1 box \\
& & & Adrenal Gland & 0.798 \cite{lee2023scaling} & 0.661 & 0.742 \cite{he2023accuracy} & 1 box \\
\cdashline{2-7}
 & \multirow{32}{*}{\makecell*[c]{\small MR}} & \multirow{9}{*}{\makecell*[c]{\small Brain}} & Brainstem & 0.860 \cite{podobnik2023han} & 0.971 & 0.692 \cite{ma2023segment} & 1 p.f, 2 p.b \\
& & & Cerebellum & 0.915 \cite{sun2023self} & 0.968 & 0.765 \cite{ma2023segment} & 1 p.f, 2 p.b \\
& & & Deep Grey Matter  & 0.974 \cite{isensee2021nnu} & 0.956 & 0.496 \cite{huang2023segment} & 1 p.f, 2 p.b \\
& & & ventricles & 0.872 \cite{lee2023d}& 0.900 & 0.639 \cite{ma2023segment} & 1 p.f, 2 p.b \\
& & & Glioma & 0.878, 0.928 \cite{ronneberger2015u} & 0.944, 0.962 & 0.763 (T1), 0.834 (FLAIR) \cite{ma2023segment} & 1 p.f, 2 p.b \\
& & & Glioma Enhancing Tumor & 0.956 \cite{lee2023scaling} & 0.952 & 0.788 \cite{ma2023segment} & 1 p.f, 2 p.b \\
& & & Glioma Tumor Core & 0.956 \cite{lee2023scaling} & 0.959 & 0.710 \cite{ma2023segment} & 1 p.f, 2 p.b \\
& & & Ischemic Stroke & 0.964 \cite{hernandez2022isles} & 0.923 & 0.613 \cite{ma2023segment} & 1 p.f, 2 p.b \\
& & & Meningioma & 0.946, 0.892 \cite{ronneberger2015u} & 0.979, 0.970 & 0.921 (T1-CE), 0.792 (T2-FLAIR) \cite{ma2023segment} & 1 p.f, 2 p.b \\
& & & Vestibular Schwannoma & 0.925 \cite{dorent2023crossmoda} & 0.952 & 0.853 \cite{ma2023segment} & 1 p.f, 2 p.b \\
& & \multirow{9}{*}{\makecell*[c]{\small Head \& Neck}} & Eye PL & 0.930 \cite{isensee2021nnu} & 0.941 & 0.815 \cite{ma2023segment} & 1 p.f, 2 p.b  \\
& & & Eye PR & 0.923 \cite{isensee2021nnu} & 0.940 & 0.819 \cite{ma2023segment} & 1 p.f, 2 p.b  \\
& & & Optic Nerve & 0.699, 0.746 \cite{isensee2021nnu} & 0.613, 0.703 & 0.395 (L), 0.433 (R) \cite{ma2023segment} & 1 p.f, 2 p.b \\
& & & Bone Mandible & 0.944 \cite{isensee2021nnu} & 0.697 & 0.543 \cite{ma2023segment} & 1 p.f, 2 p.b  \\
& & & Cricopharyngeus & 0.632 \cite{isensee2021nnu} & 0.902 & 0.614 \cite{ma2023segment} & 1 p.f, 2 p.b \\
& & & Glnd Lacrimal & 0.631, 0.621 \cite{isensee2021nnu} & 0.640, 0.687 & 0.613 (L), 0.599 (R) \cite{ma2023segment} & 1 p.f, 2 p.b \\
& & & Glnd Submand & 0.848, 0.840 \cite{isensee2021nnu} & 0.913, 0.909 & 0.779 (L), 0.797 (R) \cite{ma2023segment} & 1 p.f, 2 p.b \\
& & & Parotid & 0.871, 0.856 \cite{isensee2021nnu} & 0.917, 0.916 & 0.727 (L), 0.714 (R) \cite{ma2023segment} & 1 p.f, 2 p.b \\
& & & Glottis & 0.752 \cite{isensee2021nnu} & 0.850 & 0.301 \cite{ma2023segment} & 1 p.f, 2 p.b \\
& & & Larynx SG & 0.814 \cite{isensee2021nnu} & 0.882 & 0.540 \cite{ma2023segment} & 1 p.f, 2 p.b \\
& & & Lips & 0.722 \cite{isensee2021nnu} & 0.869 & 0.584 \cite{ma2023segment} & 1 p.f, 2 p.b \\

& & \multirow{4}{*}{\makecell*[c]{\small Abdomen}} & Left Kidney & 0.921 \cite{isensee2019automated} & 0.948 & 0.912 \cite{ma2023segment} & 1 p.f, 2 p.b  \\
& & & Right Kidney & 0.927 \cite{isensee2019automated} & 0.948 & 0.921 \cite{ma2023segment} & 1 p.f, 2 p.b \\
& & & Liver & 0.920 \cite{isensee2019automated} & 0.957 & 0.902 \cite{ma2023segment} & 1 p.f, 2 p.b \\
& & & Spleen & 0.894 \cite{isensee2019automated} & 0.948 & 0.910 \cite{ma2023segment} & 1 p.f, 2 p.b \\

& & \multirow{5}{*}{\makecell*[c]{\small Heart}} & Left Atrium & 0.933 \cite{liu2023clip} & 0.973 & 0.836 \cite{ma2023segment} & 1 p.f, 2 p.b  \\
& & & Left Ventricle & 0.959 \cite{tragakis2023fully} & 0.985 & 0.775 \cite{ma2023segment} & 1 p.f, 2 p.b \\
& & & Right Ventricle & 0.926 \cite{tragakis2023fully} & 0.972 & 0.903 \cite{ma2023segment} & 1 p.f, 2 p.b \\
& & & Artery Carotid & 0.874, 0.833 \cite{wang2023application} & 0.620, 0.627 & 0.578 (L), 0.610 (R) \cite{ma2023segment} & 1 p.f, 2 p.b \\
& & & Whole Heart & 0.867 \cite{billot2023synthseg} & 0.963 & 0.521 \cite{ma2023segment} & 1 p.f, 2 p.b \\
& & \multirow{3}{*}{\makecell*[c]{\small Prostate}} & Prostate & 0.831 \cite{isensee2021nnu} & 0.985 & 0.872 \cite{ma2023segment} & 1 p.f, 2 p.b  \\
& & & Prostate Cancer & 0.800 \cite{saha2023artificial} & 0.969 & 0.693 \cite{ma2023segment} & 1 p.f, 2 p.b \\
& & \multirow{2}{*}{\makecell*[c]{\small Spine}} & Spine & 0.952 \cite{saeed20233d} & 0.918 & 0.808 \cite{ma2023segment} & 1 p.f, 2 p.b  \\
& & & SpinalCord & 0.891 \cite{zhang2020sau} & 0.774 & 0.559 \cite{ma2023segment} & 1 p.f, 2 p.b \\

\midrule
\multirow{9}{*}{\small 2D} & \small OCT & \small Eye & Diabetic Macular Edema & 0.983 \cite{udayaraju2023combined} & 0.950 & 0.884 \cite{ma2023segment} & 1 p.f, 2 p.b \\
& \multirow{8}{*}{\makecell*[c]{\small X-Ray}} & \multirow{8}{*}{\small Chest} & Heart & 0.950 \cite{ullah2023deep} & 0.968 & 0.901 \cite{ma2023segment} & 1 p.f, 2 p.b \\
& & & Lung & 0.979 \cite{ullah2023deep} & 0.991 & 0.933 \cite{ma2023segment} & 1 p.f, 2 p.b \\
& & & Viral Pneumonia & 0.992 \cite{hasan2021deep} & 0.984 & 0.892 \cite{ma2023segment} & 1 p.f, 2 p.b \\
& & & Pneumothorax & 0.891 \cite{abedalla2021chest} & 0.815 & 0.502 \cite{ma2023segment} & 1 p.f, 2 p.b \\
& & & Tuberculosis & 0.978 \cite{liu2022automatic} & 0.969 & 0.939 \cite{ma2023segment} & 1 p.f, 2 p.b \\
& & & COVID-19 & 0.971 \cite{gholamiankhah2021automated} & 0.989 & 0.782 \cite{ma2023segment} & 1 p.f, 2 p.b \\
& & \small Breast & Breast Cancer & 0.963 \cite{alkhaleefah2022connected} & 0.833 & 0.665 \cite{ma2023segment} & 1 p.f, 2 p.b \\
\bottomrule[1.5pt]
\end{tabular}
\vspace{0.5mm}
\label{tab:rad}
\vspace{-2mm}
\end{table*}

\section{Discussion}
The continuous progression in SAM's application to medical image segmentation elucidates its potentiality in bridging the technical gap between general and domain-specific applications. As the medical imaging community gravitates towards the capabilities of SAM, it becomes paramount to reflect upon the findings, draw analytical conclusions, and understand the multifaceted challenges and requirements particular to this domain. In this section, we delve deep into the limitations of current SAM adaptation methodologies and highlight the distinctive requirements of medical image segmentation, offering a comprehensive perspective for future endeavors.
\subsection{Limitations in Current SAM Adaptation Approaches}
As we scrutinize the implementation of SAM in the realm of medical image segmentation, certain limitations inherent to the current adaptation methodologies surface:

\textbf{Generalization Discrepancy.} While SAM is adept at adjusting to a variety of tasks involving large-scale natural images, the medical imaging domain presents unique challenges, such as patient population variability and extreme pathological conditions. In addition, the trend in recent medical AI research leans towards the development of models specialized for particular applications—for instance, brain tumor segmentation or lung lesion identification—due to the highly specific nature of the knowledge required for each task. This specialization trend underscores the distinct knowledge domains in medical tasks as opposed to the more generalizable learning applicable to natural images. Thus, bridging the gap between SAM's general effectiveness with natural imagery and its potential in addressing the specialized demands of medical imaging is an unresolved issue that continues to be a focal point of research.


\textbf{Fine-tuning Dilemma.} From Table \ref{tab:T2I}, we can observe that limited recent works have been proposed to finetune the encoder network in SAM. As SAM consists of rich semantic knowledge from large-scale natural images, it is difficult to add on the medical domain knowledge with such foundation basis with limited ground-truth labeled samples. Furthermore, with the task-specific discrepancy in model training, the risk of overfitting or knowledge dilution persists and establishing an optimal strategy that balances the strengths of pre-trained weights with domain-specific adaptation remains nebulous.



\textbf{Modality Inconsistencies.} With the wide range of contrast intensity across volumetric inputs (e.g, whole images, organ-specific patches), 16 bit volumetric context are abstracted into 8-bit 2D slices (i.e. 0 to 255) as the model input and demonstrate the unify capability of performing downstream segmentation across imaging modalities and anatomical regions. However, from Table \ref{tab:rad}, we observe that the segmentation performance varies substantially and the risk of overfitting or knowledge dilution towards specific modalities may persist.



\subsection{Specific Requirement of Medical Image segmentation}
\subsubsection{Incorporation of Metadata}
Medical imaging is unique in that the raw visual data is often supplemented by a plethora of metadata, spanning patient demographics, clinical history, and technical imaging specifications. This auxiliary information, while pivotal for clinical decisions, can also enhance the richness of SAM's segmentation outcomes.

The heterogeneous nature of medical metadata presents a formidable challenge. Differences in metadata format, magnitude, and relevance across different databases and healthcare institutions necessitate the development of sophisticated normalization and preprocessing pipelines. Successfully integrating this data demands a deep understanding of the interplay between visual features and metadata, ensuring enhancements without unintentional biases. While preliminary attempts have been made in this direction, the establishment of universally-accepted, metadata-assimilation strategies harmonious with SAM's segmentation algorithms is an ongoing pursuit.

\subsubsection{Population Analysis}
The realm of medical imaging mandates a departure from an image-centric approach, propelling a shift towards population-based analyses. Here, images are interconnected fragments of larger datasets, which, when viewed in ensemble, unravel the intricate web of demographics, health trajectories, and treatment pathways.

This transition, while invaluable, is riddled with complexities. Some pathologies may predominantly affect certain demographic segments, while others may manifest diversely across age groups. Time-lapsed images from a patient can chronicle the evolution of ailments, offering insights into disease progression or therapeutic effectiveness. Ensuring SAM captures such population-driven subtleties, without being mired in the cacophony of individual variations, is a challenge warranting targeted research.

\subsection{Prospective Horizons}
As we reflect upon the current landscape of SAM's adaptation for medical image segmentation, it is evident that the journey, while marked by promising strides, is still in its nascent stages. From addressing the generalization discrepancy and fine-tuning dilemmas to navigating the intricacies of metadata incorporation and population analysis, the path forward is laden with opportunities for innovation. With such horizon, we foresee concerted efforts towards making SAM's application in medical segmentation more robust, interpretable, and clinically aligned. Here, we can summarize the potential opportunity with SAM as follows:
\subsubsection{Segmenting Unseen Classes} 
The current landscape of publicly available medical labels is heavily skewed towards specific organs or tissues, which does not adequately reflect the nuanced complexities of actual clinical environments. In response, researchers are navigating two primary paths: first, by creating datasets that represent a broader array of semantic categories, and second, by implementing incremental learning methods. Yet, both strategies rely heavily on the procurement of new annotations from a limited patient cohort for supplementary training, perpetuating a reliance on data. The advent of zero-shot learning capabilities, as exemplified by SAM, introduces a paradigm shift where human-in-the-loop prompts can potentially delineate any structure with distinct margins. Nevertheless, SAM's ability to predict based on instance-level semantics remains somewhat constrained. The emergence of CLIP has shown promising strides in aligning textual semantics with visual features, enhancing the ability to incorporate additional class information into downstream image classification tasks. Leveraging SAM's proficiency in segmenting defined regions through varied prompts, there is potential to synergize pixel-wise predictions with textual semantics. This approach could enable the classification of novel categories without the necessity of extra labeled data for training, thereby diminishing the model's dependency on extensive annotated datasets.


\subsubsection{Enhancing Explainable Interpretability}
Beyond considerations of data curation and training methodologies, SAM operates as "black boxes," its internal workings shrouded by non-linear complexities. This lack of transparency has catalyzed efforts to demystify these mechanisms, notably through the implementation of visualization techniques like feature map analysis and gradient mapping tools, such as GradCAM, during the training process. These techniques create a tangible link between the model's predictions and the features it has learned, providing a graphical exposition of how the model arrives at its conclusions. In tasks related to classification, this transparency is further enriched by examining the latent space—the realm within a network where the most complex features reside. Here, features are organized into clusters that correspond to known ground truth labels, which brings additional clarity to the model's decision-making process. The advent of prompt engineering introduces variable human-in-the-loop prompts that promise to improve the intelligibility of the relationship between feature representations and precise instance predictions. Nevertheless, parsing out singular features that correspond to multiple semantic categories, such as in multi-organ segmentation, remains a formidable challenge. Refining the relationship between feature representations and clinical outcomes, with SAM's foundational model as a starting point, represents a promising avenue toward enhancing the model's applicability in clinical settings. Such advances would contribute to the model's ability to provide more grounded, explainable results that align with clinical interpretations.

\section{Appendix}
\begin{table*}
\centering
\fontsize{7.5}{9.5}\selectfont
\caption{\textbf{Resolution of 3-D medical image segmentation datasets.} \textbf{x \& y} refers to the in-plane resolution of the pixels. \textbf{z} refers to the spacing between the slices. We show the range of the resolution for each dataset as it can vary for each scan in the dataset.}
\begin{tabular}{ccccc}
\toprule[1.5pt]
\multirow{2.5}{*}{\small \textbf{Dataset}}  & \multirow{2.5}{*}{\small \textbf{Modality}}  & \multirow{2.5}{*}{\small \textbf{Anatomy}}  & \multicolumn{2}{c}{\small \textbf{Resolution (mm)}} \\ \cmidrule{4-5}
                              &  &  &    \textbf{x \& y} & \textbf{z}  \\ \hline

BTCV~\cite{landman2015miccai}  & CT & Abdomen & 0.594-0.977 & 2.5-5.0 \\

Pancreas-CT~\cite{roth2016data,roth2015deeporgan,clark2013cancer_TCIA}  & CT & Pancreas & 0.664-0.977 & 1.5-2.5  \\

LiTS~\cite{bilic2023liver}  & CT & Liver \& Tumor  & 0.557-1.0 & 0.5-1.0 \\ 

TotalSegmentator~\cite{Wasserthal_2023_TotalSegmentator} & CT & Abdomen & \multicolumn{2}{c}{1.5 isotropic} \\

FUMPE~\cite{masoudi2018new_FUMPE}  & CT & Lung & - & $\leq$ 1.5 \\

VESSEL12~\cite{rudyanto2014comparing_VESSEL12} & CT & Lung & 0.59-0.89 & 0.5-1.0 \\

AbdomenCT-1K~\cite{abdomenct-1k} & CT & Abdomen & - & 0.5-8.0 \\

AMOS (CT)~\cite{ji2022amos}  & CT & Abdomen & - & 1.25-5.0 \\

CHAOS (CT)~\cite{CHAOSdata2019, CHAOS2021, kavur2019} & CT & Abdomen & 0.7-0.8 & 3.0-3.2 \\

HaN-Seg (CT)~\cite{podobnik2023han}& CT & Head \& Neck
 & 0.51-1.56 & 2.0-3.0 \\
 
MSD (Liver)~\cite{antonelli2022medical} & CT & Liver & 0.5-1.0 & 0.45-6.0 \\

MSD (Pancreas)~\cite{antonelli2022medical} & CT & Abdomen & - & 2.5 \\

\hline

ACDC~\cite{ACDCData}  & MRI & Heart & 1.170-1.296 & 5.0-10.0 \\

M\&Ms~\cite{MMsData}  & MRI & Heart &  0.684-1.823 & 5.0-10.0 \\

BraTS~\cite{karargyris2023federated, baid2021rsna, menze2014multimodal, bakas10segmentation, bakas2017advancing, bakas2017segmentation, labella2023asnr, moawad2023brain, kazerooni2023brain, adewole2023brain}  & MRI & Brain
 &  \multicolumn{2}{c}{1.0 isotropic} \\

NCI-ISBI~\cite{Bloch2015_NCI-ISBI2013}  & MRI & Prostate
 & - & 3.0-4.0 \\

PROMISE12~\cite{Promise12Data} & MRI & Prostate & 0.273-0.750, & 2.2-4.0 \\

AMOS (MRI)~\cite{ji2022amos}  & MRI & Abdomen & - & 0.82-6.0 \\

CHAOS (MRI)~\cite{CHAOSdata2019, CHAOS2021, kavur2019} & MRI & Abdomen & 1.36-1.89 & 5.5-9.0 \\

HaN-Seg (MRI)~\cite{podobnik2023han}& MRI & Head and Neck
 & 0.47-0.82 & 3.0-5.0 \\

MSD (Brain)~\cite{antonelli2022medical} & MRI & Brain & \multicolumn{2}{c}{1.0 isotropic} \\

MSD (Hippocampus)~\cite{antonelli2022medical} & MRI & Brain & \multicolumn{2}{c}{1.0 isotropic} \\

MSD (Heart)~\cite{antonelli2022medical} & MRI & Heart & 1.25 & 2.7 \\

MSD (Prostate)~\cite{antonelli2022medical} & MRI & Prostate & 0.6 & 4.0 \\

\bottomrule[1.5pt]
\end{tabular}
\label{tab:3d-dataset}
\vspace{-2mm}
\end{table*}

\begin{table*}[]
\centering
\fontsize{7}{9.6}\selectfont
\caption{\textbf{Quantitative Comparisons between the Zero-Shot Performance of SAM and Current State-Of-The-Art (SOTA) Approaches on Different Pathology Datasets.}}
\begin{tabular}{ccccccc}
\toprule[1.5pt]
\multirow{2}{*}{\small Modality} & \multirow{2}{*}{\small Region} & \multirow{2}{*}{\small Tasks} & \multirow{2}{*}{\small Resolution} & \multicolumn{2}{c}{\small Performance} & \multirow{2}{*}{\small Prompt Mode}\\
& & & & \small SOTAs & \small SAM & \\
\midrule
\multirow{9}{*}{\small H \& E} & \small Skin & Tumor & 0.5 $\times$ & 0.720 \cite{deng2023segment} & 0.750 & 20 points \\
& \multirow{6}{*}{\small Kidney} & Glomerular (CAP) & \multirow{2}{*}{5 $\times$} & 0.965 \cite{deng2023segment} & 0.801 & 20 points \\
& & Glomerular Tuft (TUFT) & & 0.966 \cite{deng2023segment} & 0.799 & 20 points \\
& & Distal Tubular (DT) & \multirow{2}{*}{10 $\times$} & 0.810 \cite{deng2023segment} & 0.604 & 20 points \\
& & Proximal Tubular (PT) & & 0.898 \cite{deng2023segment} & 0.666 & 20 points \\
& & Arteries (VES) & & 0.851 \cite{deng2023segment} & 0.685 & 20 points \\
& & Peritubular Capillaries (PTC) & \multirow{3}{*}{40 $\times$} & 0.772 \cite{deng2023segment} & 0.646 & 20 points \\
& \small Different Tumors & Nuclei & & 0.818 \cite{deng2023segment} & 0.417 & 20 points \\
& \small Colon & Adenocarcinoma, Benign Glands & 20 $\times$ & 0.797 \cite{wang2021hybrid} & 0.525 & classes prompt \\

\bottomrule[1.5pt]
\end{tabular}
\vspace{0.5mm}
\label{tab:path}
\vspace{-2mm}
\end{table*}

\begin{table*}[]
\centering
\fontsize{7}{9.6}\selectfont
\caption{\textbf{Quantitative Comparisons between the Zero-Shot Performance of SAM and Current State-Of-The-Art (SOTA) Approaches on Different Camera Datasets.}}
\begin{tabular}{cccccc}
\toprule[1.5pt]
\multirow{2}{*}{\small Modality} & \multirow{2}{*}{\small Region} & \multirow{2}{*}{\small Tasks} & \multicolumn{2}{c}{\small Performance} & \multirow{2}{*}{\small Prompt Mode}\\
& & & \small SOTAs & \small SAM & \\
\midrule
\multirow{4}{*}{\small Endoscopy} & \multirow{3}{*}{\small Abdomen} & Cholecystectomy & 0.955 \cite{ma2023segment} & 0.718 & 1 p.f, 2 p.b \\
&  & Instruments & 0.940 \cite{he2023accuracy} & 0.893 & 1 box \\
&  & Gastrectomy & 0.974 \cite{ma2023segment} & 0.815 & 1 p.f, 2 p.b \\
& \small Colon & Polyp & 0.925 \cite{he2023accuracy} & 0.872 & 1 box \\
\small Dermoscopy & \small Skin & Cancer & 0.906 \cite{he2023accuracy} & 0.865 & 1 box \\
\multirow{2}{*}{\small Fundus} & \multirow{2}{*}{\small Retina} & Optic Disc & 0.983 \cite{wu2023medical} & 0.924 & 1 box \\
& & Fundus Vasculature & 0.809 \cite{qiu2023learnable} & 0.217 & 1 point \\

\bottomrule[1.5pt]
\end{tabular}
\vspace{0.5mm}
\label{tab:cam}
\vspace{-2mm}
\end{table*}

{\small
\bibliographystyle{ieee_fullname}
\bibliography{mybib}
}

\end{document}